\documentclass[journal]{IEEEtran}

\ifCLASSINFOpdf

\else

\fi

\usepackage{tabularx,ragged2e,booktabs,caption}
\usepackage{booktabs}
\usepackage{siunitx}
\usepackage{graphicx}
\usepackage{caption}
\usepackage{amsmath, amsthm, amssymb, amsfonts}

\begin{document}

\title{A Review of Automated Pain Assessment in Infants: \\ Features, Classification Tasks, and Databases}

\author{Ghada~Zamzmi,~\IEEEmembership{Member,~IEEE,}
Ruicong~Zhi,~\IEEEmembership{Member,~IEEE,}
Rangachar~Kasturi,~\IEEEmembership{Fellow,~IEEE,}
Dmitry~Goldgof,~\IEEEmembership{Fellow,~IEEE,}
Terri~Ashmeade,~\IEEEmembership{MD,}
and~Yu~Sun,~\IEEEmembership{Member,~IEEE}
\IEEEcompsocitemizethanks{\IEEEcompsocthanksitem The authors, except Ruicong Zhi, are with Computer Science and Engineering Department (Ghada Zamzmi, Rangachar Kasturi, Dmitry Goldgof, and Yu Sun) and Pediatrics Department (Terri Ashmeade), University of South Florida, Tampa,
FL, 33620. \protect \\
Ruicong Zhi is with the School of Computer and Communication Engineering, University of Science and Technology Beijing; Beijing Key Laboratory of Knowledge Engineering for Materials Science, Beijing 100083, P.R.China \protect \\
(E-mail:\{ghadh,r1k,goldgof,tashmead,yusun\}@mail.usf.edu; zhirc@ustb.edu.cn)
 }}

\markboth{IEEE REVIEWS IN BIOMEDICAL ENGINEERING}%
{Zamzmi\MakeLowercase{\textit{et al.}}: Automated Pain Assessment in Infants}

\maketitle

\begin{abstract}
Bedside caregivers assess infants' pain at constant intervals by observing specific behavioral and physiological signs of pain. This standard has two main limitations. The first limitation is the intermittent assessment of pain, which might lead to missing pain when the infants are left unattended. Second, it is inconsistent since it depends on the observer's subjective judgment and differs between observers. The intermittent and inconsistent assessment can induce poor treatment and, therefore, cause serious long-term consequences. To mitigate these limitations, the current standard can be augmented by an automated system that monitors infants continuously and provides quantitative and consistent assessment of pain. Several automated methods have been introduced to assess infants' pain automatically based on analysis of behavioral or physiological pain indicators. This paper comprehensively reviews the automated approaches (i.e., approaches to feature extraction) for analyzing infants' pain and the current efforts in automatic pain recognition. In addition, it reviews the databases available to the research community and discusses the current limitations of the automated pain assessment.
\end{abstract}


\IEEEpeerreviewmaketitle

\section{Introduction}
\IEEEPARstart{P}{hysical} pain can be defined as the negatively-valenced experience associated with actual or potential tissue damage \cite{loeser2008kyoto}. Pain in neonates can be categorized into two main types: acute procedural pain and acute prolonged pain \cite{anand2007pain}. Acute procedural pain is often caused by a short painful stimulus (e.g., immunization) and it ends as soon as the cause of pain (i.e., stimulus) is removed. The acute prolonged pain (a.k.a., postoperative) is triggered by a clear stimulus (e.g., surgical procedure) and has a clearly defined beginning and expected end point; the intensity of this type of pain decreases as a function of time since the painful stimulus occurred. Infants may experience different types of pain simultaneously.

The accurate assessment of pain is vital because it helps caregivers understand the severity of the patient's situation and develop appropriate treatments. The most well-known pain assessment method is the patient's self-evaluation. Another common pain assessment method is the Visual Analog Scale (VAS) that has symbols or numbers to denote different levels of pain. Although these methods are the gold standards for clinical assessment, they are not applicable for infants.

The current standard for assessing pain in this vulnerable population depends on the caregivers' observation of specific behavioral (e.g., facial expression) and physiological (e.g., vital signs) pain indicators. Table I summarizes the most common pediatric pain scales for different types of pain. As stated in \cite{mcgrath2013oxford}, most of the existing pain scales are designed for procedural pain and a few are designed for prolonged pain. The interested reader is referred to \cite{anand2007pain,macdonald2012atlas,mcgrath2013oxford} for more information about the validity and shortcomings of different neonatal pain scales. 

Assessing infants' pain manually using the common pediatric scales has three limitations. First, caregivers assess pain at different time intervals and are not able to provide continuous assessment of pain. Continuous monitoring is important because infants might experience pain when they are left unattended. This is especially true for postoperative pain since it requires continuous intensive care and prompts pain detection and intervention. Second, caregivers' assessment of pain is highly biased and is affected by several idiosyncratic factors, such as the observer's cognitive bias, identity \cite{pillai2004parental,samolsky2016medical}, background and culture \cite{pillai2004parental,pillai2006judgments,pillai2008understanding}, and gender \cite{miller2006pain}, that may lead to inconsistent assessment and treatment of pain. Third, the current practice for assessing infants' pain is time-consuming and requires a large number of trained and professional labors, which makes it infeasible in low-income countries where the medical professionals and resources are scarce.

\begin{table*}
\caption{Examples of Common Pediatric Pain Scales} 
\label{table:nonlin}
\begin{tabularx}{\textwidth}{*{6}{l}} 
\toprule
\textbf {Pain Scale} & \textbf{Pain Type} & \textbf {Age Range} & \textbf{Behavioral Measures} & \textbf{Physiological Measures} & \textbf{Psychometric Properties \footnotemark}  \\
\midrule
\\
\cite{lawrence1993development} Neonatal & Procedural & 28-38 & Facial expression, &  Breathing patterns & Inter-rater reliability: \\ 
Infant Pain Scale  &  & gestations weeks & crying, arms/legs, &  & (r=0.92-0.97)\\ 
(NIPS) &  &  & movement, and &  & Internal Consistency: \\ 
  &  &  & arousal state &  & (Cronbach's $\alpha$= 0.87-0.95)  \\ 
  &  &  &  &  & Content validity   \\ 
    &  &  &  &  & Concurrent validity:   \\ 
    &  &  &  &  & (r=0.53-0.83)   \\ \addlinespace
\hline
\\
\cite{GRUNAU1987395} Neonatal & Procedural & Preterm $\geq$ 25 & Brow bulge, &  NA  & Inter-rater and Intra- \\
Facial Coding  &  & gestations weeks & Eye squeeze &  & rater reliability $\ge$ 0.85 \\ 
System (NFCS) &  & to term infants  & Nasolabial furrow, &  & Internal Consistency: \\
  &  &  & Open lips, &  & (Cronbach's $\alpha$ = 0.87-0.95)  \\ 
  &  &  & Horizontal mouth,  &  & Content and face validity   \\ 
    &  &  & Vertical mouth,  &  & Construct validity    \\ 
    &  &  & Lips pursed,  &  &  \\ 
    &  &  & Taut tongue,  &  &   \\ 
    &  &  & Chin quiver,  &  &   \\ 
    &  &  & Tongue protrusion  &  &   \\ \addlinespace
\hline
\\
\cite{hummel2008clinical} Neonatal & Postoperative & 23-40 & Facial expression, &  Heart rate, & Inter-rater reliability:  \\ 
Pain, Agitation,  &  & gestations weeks & behavior movements, & respiratory rate, & (r=0.85-0.95) \\ 
and Sedation Scale  &  &   & crying/irritability, & blood pressure, & Intra-rater reliability: \\ 
(N-PASS)  &  &  & and extremities tone & and oxygen saturation & (r$=$0.87)  \\ 
  &  &  &   &  & Internal consistency:    \\
    &  &  &  &  & (Cronbach's $\alpha$ $=$ 0.84-0.89)    \\ 
    &  &  &  &  & Construct validity: (P $\le$ .0001)    \\ \addlinespace
\hline
\\
\cite{krechel1995cries} Crying, & Postoperative & 32 -- 60 & Facial expression, & Requires increased & Inter-rater reliability:  \\
Requires O2, &  & gestations weeks & crying, and, & oxygen and VS & (r$=$0.98) \\ 
Increased VS,  &  &   & sleeping state &  & Construct and content\\ 
Expression, and &  &  &  &  & validity  \\ 
Sleepless (CRIES)  &  &  &   &  &    \\
\hline
\end{tabularx}
\end{table*}

The intermittent and inconsistent assessment of pain might lead to misdiagnosis and over/under treatment. Different pediatric studies \cite{zwicker2016smaller,vinall2012neonatal,american2007prevention,page2004there} reported that the inadequate pain treatment is associated with an increase in the avoidance behaviors and social hypervigilance and it can cause long-term changes in the brain structure (e.g., cause alterations in the cerebral white matter and subcortical grey matter). These alterations can lead to a variety of behavioral, developmental, and learning disabilities \cite{page2004there}. Consequently, developing automated systems that provide continuous and more consistent pain assessment is important.

In the past several years, there has been an increasing interest in the use of machine-learning methods for understanding human behavioral responses to pain based on analysis of facial expressions (\cite{4042206,Hammal:2012:ADP:2388676.2388688,HAMMAL20121265,lucey2008improving,ASHRAF20091788,5771462,wei2011pain,912537,6553762,Sikka2014,zhu2014pain,niese2009towards,LITTLEWORT20091797,BARTLETT2014738,werner2014comparative,florea2014learning,werner2013towards,kachele2015multimodal,kachele2017adaptive,6977497,6617456,7349412,velana2016senseemotion,Kaltwang2012,Pal1660444,ROY201699,martinez2017338,RATHEE201677,6607608,RATHEE2015247,brahnam2005svm,BRAHNAM2006211,BRAHNAM20071242,5415598,nanni2010local,zamzami2015pain,Fotiadou2014,sikka2015automated,liu2017deepfacelift}), crying sound (\cite{Pal1660444,pai2016automatic,Varallyay1403155,SCHEINER2002509,Petroni205186,6176851,FULLER1988251}), and body movement (\cite{7900284,Zamzmi2017}). Also, studies have shown that automated systems can be used to detect emotions from physiological responses such as pupil dilation (\cite{Partala2000,PARTALA2003185,6038772, Al-Omar2013}), galvanic skin response (GSR) (\cite{6977497,kachele2015multimodal,gruss2015pain,Yoo2005}), changes in heart rate (\cite{Yoo2005,faye2010newborn,WALTER2014,gruss2015pain}), and cerebral hemodynamic changes (\cite{brown2011towards,RANGER2014519}). A short review of the current efforts in analyzing pain emotion automatically and a discussion of challenges is presented in \cite{Hammal_2014}. 

\footnotetext{Properties to define instruments' reliability (i.e., consistency) and validity (i.e., accuracy).}

In this review, we extensively and specifically explore the current efforts for assessing infants' pain automatically. The main contributions of this paper can be summarized as follows: 

 \begin{itemize}
  \item We present a structured review of the current methods for extracting pain-relevant features from infants' data (Section II). We divided these methods into three main categories, behavioral-based, physiological-based, and multimodal-based. These categories were divided further, based on the utilized pain indicator, into facial expression, body movement, crying sound, vital signs, and cerebral hemodynamic. Then, each of these categories was divided further as illustrated in Figure 1. 
 \item We propose to categorize the existing pain recognition works into pain detection and pain intensity estimation. We define pain detection as the task of detecting the presence or absence of pain and pain intensity estimation as the task of estimating the intensity of the detected pain (i.e., how much an infant is in pain). Description and a discussion of limitations for each classification task is presented in Section III. 
\item We review the pain databases that are available for research use (Sections IV), discuss the current limitations of automated pain assessment, and suggest directions for future research (Section V). 
\end{itemize}

Before we proceed, we would like to note that this review does not discuss preprocessing operations (e.g., image or signal enhancement, noise reduction, region of interest detection, facial landmark detection, etc.) since they are beyond the paper's scope. We refer the interested reader to \cite{kaur2011survey,bedi2013various} for a review of image enhancement methods, to \cite{kaur2012comparative} for signal denoising methods, to \cite{854761,6248014} for region of interest detection, and to \cite{sagonas2013300,Zhang2014} for facial landmark detection. In addition, we note that understanding this paper requires basic knowledge of machine-learning concepts such as feature (i.e., a measurable property of an object), feature vector (i.e., n-dimensional vector of numerical features), classifier's accuracy, and other performance evaluation techniques. A simple, yet comprehensive, explanation of these concepts can be found in \cite{witten2016data,deo2015machine}.

\textbf{Organization:} Section II presents the current methods that analyze pain automatically to extract pain relevant features. Section III discusses the current state-of-art for pain recognition. Section IV provides summary of pain databases that are available for research use. We list several challenges and discuss future directions of pain assessment in Section V. Finally, we conclude in Section VI. 

\section{Pain Analysis}
The automated analysis of infants' pain is an emerging topic in artificial intelligence due to the increasing demands for continuous and consistent monitoring of pain in clinical environments and homecare. Numerous methods have been introduced to automatically detect infants' pain based on analysis of behavioral or physiological pain indicators or a combination of both. We grouped these methods into three main categories, namely behavioral measures based pain analysis, physiological measures based pain analysis, and multimodal based pain analysis, and divided these categories further as illustrated in Figure 1.

\subsection{Behavioral Measures Based Pain Analysis}
Behavioral measures based pain analysis can be defined as the task of automatically extracting pain-relevant features from behavioral pain indicators such as facial expression and crying sound. In this section, we discuss the existing methods that analyze facial expression, crying, or body movement to extract useful features for classification.  

\subsubsection{Facial Expression}
Facial expression is one of the most common and specific indicators of pain. Facial expression of pain is defined as the movements and distortions in facial muscles associated with a painful stimulus. The facial movements associated with pain in infants include deepening of the nasolabial furrow, brow lowering, narrowed eyes, vertical and horizontal mouth stretch, lip pursing, lip opening, tongue protrusion, taut tongue, and chin quiver \cite{GRUNAU1987395}.

Automatic recognition of pain expression consists of three main stages: (1) face detection and registration; (2) feature extraction; and (3) expression recognition (see Section III). Face detection is a mature area of research and, therefore, will not be discussed further. Several methods have been proposed to extract pain-relevant features from images. We broadly divided these methods based on their underlying algorithms into five groups: Feature Reduction Based methods, Local Binary Pattern Variation based methods, Motion-based methods, Model-based methods, and Facial Action Coding System [FACS] (see Figure 1). The first and second categories focus on analyzing static images and they both fall under texture-based methods. The last three categories focus on the temporal analysis of facial expression in videos. For each category, we discuss the underlying algorithms and the exiting works that utilize them. Table II presents a summary of the works we discussed in this section.

\begin{figure}[!t]
\centering
\vspace{0 cm}
\includegraphics[scale=0.21]{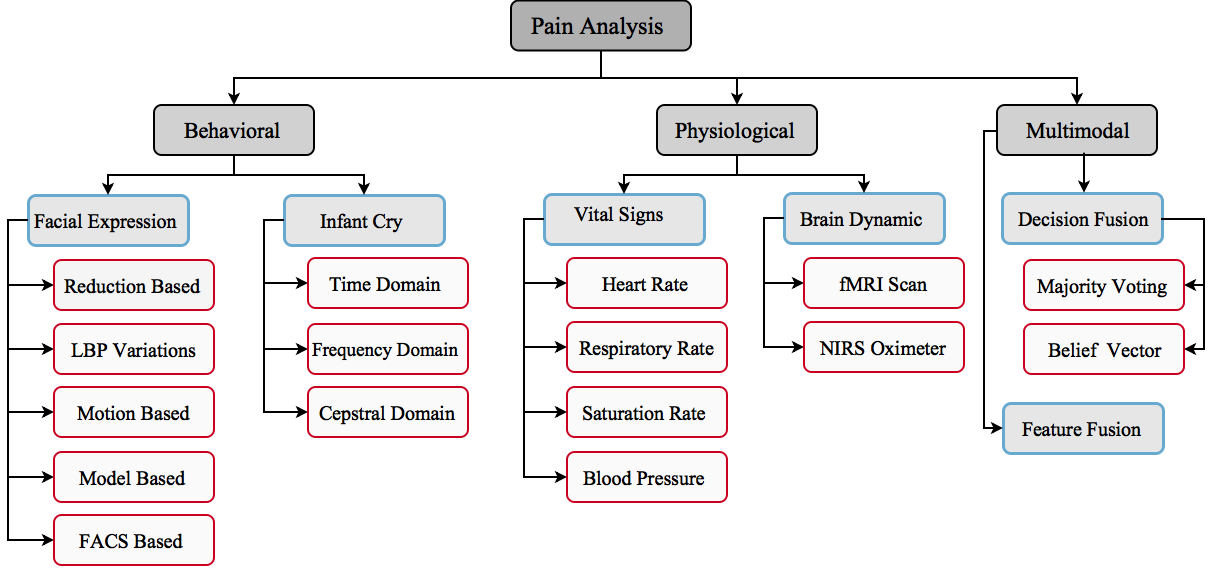}
\centering
\caption{Tree Diagram of Infants' Pain Analysis Methods. }
\end{figure}

\begin{table*}
\caption{Summary of Automated Methods for Analyzing Infants' Pain Expression} 
\label{table:nonlin}
\begin{tabularx}{\textwidth}{*{6}{l}} 
\textbf {Ref. \& Year} & \textbf{Database} &\textbf{Category} & \textbf {Extraction Method} & \textbf{Classification} & \textbf{Results} \\
\addlinespace
\cite{BRAHNAM2006211}: 2006 & COPE Database & Feature & Column stacking & PCA/LDA with L1 & SVM avg. accuracy:\\ 
 & \textbf{Subjects:} 26, half girls  & Reduction & image's intensities & and SVM: Pain/nopain, & Pain/nopain (88\%)\\ 
  & \textbf{Race:} Caucasian  & Based & and dimensionality &  pain/rest, pain/cry  & Pain/rest (95\%) \\ 
 & \textbf{Age range:} 18 hours to 3 days&  & reduction (PCA) & pain/air-puff, and & Pain/cry (80\%)  \\
  &  \textbf{Stimuli:} Pain stimulus and 3 &  & & pain/friction & Pain/air-puff (83\%) \\ 
  & other stimuli: air, friction, &  &  & \textbf{Testing Protocol:} & Pain/friction (93\%)\\ 
    & and rest/cry &  &  & 10-fold cross-validation & 
 \\ 
    &\textbf{Data:} 204 static images &  &   & \\ \addlinespace
\hline
\\
\cite{BRAHNAM20071242}: 2007 & COPE Database & Feature & Column stacking &  NNSOA, PCA/LDA & Average accuracy: \\ 
  &  & Reduction & image's intensities & and SVM: & NNSOA (90.20\%) \\ 
 &  & Based  & and dimensionality & Pain (60 images) vs & SVM (82.35\%) \\ 
  &  &  & reduction & no pain (144 images) & PCA w/ L1 (80.35\%) \\ 
  &  &  &  & \textbf{Testing Protocol:} &  LDA w/L1 (76.96\%) \\ 
    &  &  &   & Leave-one-subject-out &    \\
    &  &  &  & cross-validation  &   \\ \addlinespace
\hline
\\

\cite{5415598}: 2010 & COPE Database & Feature & Column stacking & RVM:  & Weighted Kappa  \\ 
  &  & Reduction  &  image's intensities & Pain/nopain  &  Coeff.: \\ 
  &  & Based &  & Pain Intensity & 0.47 (expert/RVM) \\ 
 &  &  &  & Estimation & 0.46 (non-expert/RVM) \\ 
  &  & &  & \textbf{Testing Protocol:} &    \\ 
    &  &  &  & Leave-one-image-out  &    \\
    &  &  &  & cross-validation  &     \\ \addlinespace
\hline
\\
\cite{nanni2010local}: 2010 & COPE Database  & LBP Variation & LBP, LTP, ELTP, & SFFS feature selection  &  Highest (0.93) area \\ 
 &  & Based & and ELBP & and SVMs & under the curve of  \\ 
  &  &  & descriptors & \textbf{Testing Protocol:} & ROC (ELTP) \\ 
  &  &  &  & Leave-one-out & \\ 
  &  &  &   & cross-validation (SFFS) &     \\ 
  &  &  &   & Train/Test Split  &     \\ \addlinespace
  \hline
\\
\cite{zamzami2015pain}: 2015  & \textbf{Subjects:} 10, half girls & Motion & Strain magnitude & KNN and SVM: & Highest overall \\ 
 & \textbf{Race:} Caucasian  & Based & estimated from & Pain vs  no pain & accuracy: \\ 
  &\textbf{Age range:} 32 to 41  & (Optical Flow) & flow vectors  &  expression & KNN, (96\%) \\ 
 & gestations weeks&  &  &  \textbf{Testing Protocol:} &  \\ 
  &  \textbf{Stimuli:} Pain stimulus  &  & & 10-fold cross-validation &  \\ 
  & (i.e., heel lancing) and  &  &  &  & \\ 
    & normal state &  &  &  & 
 \\ 
    &\textbf{Data:} 10 videos and NIPS scores &  &   & \\
        \addlinespace
 \hline
 \\
\midrule

\cite{Fotiadou2014}: 2014  & \textbf{Subjects:} 10 & Model & AAM-based & SVM: & AUC of ROC: \\ 
& \textbf{Race:} N/A  & Based & features & Discomfort vs  & 0.98 \\ 
  &\textbf{Age range:} N/A  & (AAM) & SPTS, SAPP,   &  comfort &  \\ 
 & \textbf{Stimuli:} Heel puncture, diaper &  & and CAPP &  \textbf{Testing Protocol:} &  \\ 
  &  change, hunger, and resting  &  & & Leave-one-subject-out &  \\ 
 & \textbf{Data:} 15 videos, ranges from &  &  & cross-validation & \\ 
   & few seconds to minutes &  &  &  & \\ 
       \addlinespace
        
         \hline
\\

\cite{sikka2015automated}: 2015  & \textbf{Subjects:} 50 children, 35\% boys & FACS & Strain magnitude & KNN and SVM: & Highest overall \\ 
 & \textbf{Race:} 35 Hispanic, 9 non-  & Based & estimated from & Pain expression vs  & accuracy: \\ 
  & Hispanic white, 5 Asian, and   & (Optical Flow) & flow vectors  &  no pain expression & KNN, (96\%) \\ 
 & 1 Native American &  &  &  \textbf{Testing Protocol:} &  \\ 
  &  \textbf{Age range:} 5 to 18 years  &  & & 10-fold cross-validation &  \\ 
  & \textbf{Stimuli:} appendectomy (ongoing)  &  &  &  & \\ 
    & and pressing surgical site (transient) &  &  &  & 
 \\ 
    & \textbf{Data:} videos, self-report, and by  &   & \\
        & proxy rating by a nurse and parent &  &   & \\
\addlinespace 
\\
\hline
\end{tabularx}
\end{table*}

\paragraph{\textbf{Feature Reduction Based Methods}}
A simple approach to extract pain-relevant features from static images is to convert the image's pixels into a vector of $N_x \times N_y \times 1$ dimensions, where $N_x$ and $N_y$ represent the image's width and height. Then, feature reduction methods such as Principal Component Analysis (PCA) and Sequential Floating Forward Selection (SFFS) could be applied to reduce the vector's dimensionality.  

PCA is a statistical method to reduce the dimensionality of a given feature space by identifying a small number of uncorrelated features or variables, known as principle components. Those components represent the dimensions along which the data points are mostly spread out. A detailed explanation of PCA along with its mathematical formulation can be found in \cite{wold1987principal}. 

Another well-known method for feature selection is SFFS \cite{VERVERIDIS20082956}. Sequential feature selection methods are a family of greedy search algorithms that are used to reduce an initial d-dimensional feature space to a k-dimensional feature subspace where $k < d$ by sequentially adding or removing a single feature until there is no improvement in the classifier performance. SFFS is an extension of Sequential Forward Selection (SFS), which is a method to construct the best feature subset by adding to a subset, initially equal to null, a single feature that satisfies some criterion function. The difference between SFS and SFFS is that SFFS allows, according to the criterion function, to exclude the worst feature from the subset (i.e., allow to dynamically increase and decrease the features until the best subset is reached). 

One of the first studies in machine recognition of pain is presented by Brahnam et al. in \cite{brahnam2005svm,BRAHNAM2006211}. A feature reduction based approach was proposed \cite{BRAHNAM2006211} and applied on the Classification of Pain Expressions (COPE) dataset. This dataset consists of 204 color images captured for 26 Caucasian infants, half girls, using Nikon D100 digital camera. The infants' age ranges from eighteen hours to three days old and all infants were in good health. The face images of infants were taken while experiencing four different stimuli: pain stimulus during the heel lancing (60 images), rest/cry stimulus during the transportation of an infant from one crib to another (63 rest images and 18 cry images), air stimulus to the nose (23 images), and the friction stimulus, which involves receiving friction on the external lateral surface of the heel with cotton soaked in alcohol (36 images). To extract pain-relevant features, each image was rotated, cropped, converted to grayscale, and reduced to $ 100 \times 120 $ pixels. The rescaled image was then concatenated into a feature vector of 12000 dimensions with values ranging from 0 to 255. To reduce the high dimensionality of this vector, PCA was applied. 

For classification, distance-based classifiers, specifically PCA and Linear Discriminant Analysis (LDA), and Support Vector Machine (SVM) were used to classify the infants' images into one of the following pairs: pain/no-pain, pain/rest, pain/cry, pain/air puff, and pain/friction. LDA can be defined as a supervised learning method that works by transforming the data (i.e., images) onto a subspace that maximizes the ratio of “between-class variance” to “within-class variance” in order to increase the separation between classes. SVM is another supervised machine learning algorithm that works by constructing the optimal separating hyperplane that best segregates new data or examples. The results showed that SVM with a polynomial kernel of degree 3 evaluated using 10-fold cross-validation model achieved the best recognition rate and outperformed distance-based classifiers in classifying pain versus no-pain (88.00\%), pain versus rest (94.62\%), pain versus cry (80.00\%), pain versus air-puff (83.33\%), and pain versus friction (93.00\%). 

The above-discussed work is extended in \cite{BRAHNAM20071242} to include Neural Network Simultaneous Optimization Algorithm (NNSOA) for classification along with LDA, PCA, and SVM. Moreover, leave-one-subject-out cross-validation for a total of 26 subjects was used to evaluate NNSOA classifier instead of 10-fold cross-validation; It has been reported \cite{Brahnam2007} that leave-one-subject-out cross-validation protocol is challenging but more realistic in clinical settings. The results showed that NNSOA has the highest average classification rate (90.20\%) in classifying infants' images as pain (60 images) or no pain (144 images). SVM, PCA and LDA with L1 distance achieved average classification rates of 82.35\%, 80.39\% and 76.96\%, respectively.

Instead of detecting the presence or absence of the pain expression, Golahmi et al. \cite{5415598} presented a \textit{sparse} kernel machine algorithm, known as Relevance Vector Machine (RVM), to estimate the intensity level of the detected pain expression. RVM is a Bayesian version of SVM that provides the posterior probabilities for the class memberships. In the preprocessing stage, a total of 181 images from COPE dataset were standardized using similarity transformation, cropped using $ 70 \times 93 $ window to get the exact facial region, and converted to grayscale. Then, each image was converted into a 6510-dimensional vector by column stacking the intensity values. These vectors were used to build RVM model and leave-one-image-out method was applied for the model's evaluation. 

For system's validation, the estimated pain intensity generated by the RVM algorithm (i.e., posterior probability or uncertainty of the class membership) was compared with pain intensity assessment of five expert and five non-expert examiners. To give the human examiners a prior knowledge for the assessment, two images, corresponding to pain and no-pain conditions, were selected for each infant and assigned a score of 0 (i.e., no-pain) and 100 (i.e., pain). The human examiners were then asked to provide a multiple of ten score that ranges from 0 to 100 for each image. The results showed moderate agreement between human expert examiners assessment (0.47 Weighted Kappa Coefficient with 95\% confidence interval of 0.37 to 0.57) and non-expert examiners assessment (0.46 Weighted Kappa Coefficient with 95\% confidence interval of 0.36 to 0.55) as compared with the RVM assessment. The agreement, measured using the Weighted Kappa Coefficient, between human experts and human non-experts was 0.78 with a 95\% confidence interval of 0.73 to 0.82.

\paragraph{\textbf{Local Binary Pattern Variation Based Methods}} The methods presented in this category utilize Local Binary Pattern (LBP) algorithm or its variants for analysis. Local Binary Pattern (LBP) is one of the most popular texture descriptors in computer vision. The popularity of LBP can be attributed to its simplicity, its low computational complexity, and its robustness to illumination variations and alignment error \cite{Ren2016}.

The basic LBP \cite{ojala2002multiresolution} describes the image's texture by comparing the gray value of a central pixel $X$ with the gray values of its $P$ neighbors within a predefined circle of radius $R$ and considering the output of the comparison as a binary number. For example, the value of a neighbor pixel would be one if the value of that pixel is greater than the central pixel value and zero otherwise. These binary values are then encoded to form local binary patterns, converted into decimal, and accumulated into a discrete histogram. The original LBP is not rotation invariant. Therefore, an extension was proposed to make it rotation invariant by performing $P$ bitwise shift operations on the binary patterns and choosing the smallest value as the output. Another extension of LBP, known as Improved LBP, that is less sensitive to noise is presented in \cite{jin2004face}. This descriptor reduces the image's noise by comparing the intensity of the neighboring pixels with the local mean value instead of the central pixel.

Local Ternary Pattern (LTP) \cite{tan2010enhanced} is another extension of LBP. The main difference between LBP and LTP is that the difference between the central pixel $X$ and its neighbors $P$ is represented by a 3-valued function \cite{tan2010enhanced}. An Elongated Binary Pattern (ELBP) \cite{liao2007face} and Elongated Ternary Pattern (ELTP) \cite{nanni2010local} are variants of LBP and LTP that use an elliptic neighborhood window instead of a circular window. As discussed in \cite{liao2007face}, the elliptic neighborhood window allows to capture the anisotropic structure of facial images more effectively. 

Nanni et al. \cite{nanni2010local,nanni2010localm} presented a method to detect infants' expressions of pain using LBP, LTB, ELTP, and ELBP texture descriptors. These descriptors were applied on COPE dataset (i.e., 204 static images photographed during pain-inducing event and other obnoxious stimuli) to extract pain-relevant features. In the preprocessing stage, the images were re-sized, aligned, cropped to obtain the exact facial region, and divided into blocks or cells of $25 \times 25$ dimensions. To select the most discriminate cells, SFFS feature selection algorithm was applied to a training set using leave-one-out cross-validation evaluation protocol. 

For classification of infants' images as pain or no-pain, an ensemble of Radial Basis SVMs was built and evaluated on a testing set. The study's results showed that ELTP texture descriptor achieved the highest (approx. 0.93) Area Under the Curve of Receiver Operating Characteristic curve (AUC of ROC) as compared to other texture descriptors. It also showed that pain expression affect sub regions of the face and thus dividing the whole image into cells can improve the performance.

A noticeable limitation of this work is the use of static texture descriptors for detecting facial expressions. Static texture descriptors deal only with the spatial information and ignore the dynamic pattern of facial expressions. To measure the spatiotemporal information of facial expressions, several dynamic texture descriptors such as Local Binary Patterns on Three Orthogonal Planes (LBP-TOP) \cite{zhao2007dynamic} can be exploited. 

There is a limitation in using 2D static images (COPE dataset) for pain expression recognition. Static images ignore the expression's dynamic and temporal information. It effects on the ability of understanding the facial expression and its evolution over time. Occlusion (e.g., self-occlusion, oxygen mask, and pacifier), which is known to be common in clinical environments, is another limitation of using static images. We present below several methods to analyze pain expression dynamically from videos. 

\paragraph{\textbf{Motion-based Methods}}
Motion-based methods can be defined as the methods that estimate the motion vectors for a pixel (direct) or features (indirect) between consecutive video frames. Optical flow is a well-known motion estimation method that works by directly estimating the pixel's velocity over consecutive video frames. It depends on the brightness conservation principle and provides a dense pixel-to-pixel correspondence. More discussion about optical flow algorithm and its implementation can be found in \cite{correia2002real}. 

Zamzmi et al. \cite{zamzami2015pain} introduced a motion-based method to detect infants' pain expression from videos. The dataset utilized in this work was collected from 10 infants, age ranges from 32 to 41 gestations weeks, hospitalized in the Neonatal Intensive Care Unit (NICU) at Tampa General Hospital (TGH). The videos were recorded for infants undergoing routine acute painful procedure (i.e., heel lancing). Specifically, the infants were recorded prior the painful procedure in normal state to get the baseline. Then, they were recorded during the painful procedure (i.e., from the start till the end of the procedure) and after the completion of the painful procedure. To get the ground truth labels, trained nurses were asked to assess infants' pain and provide scores using a pediatric pain scale known as NIPS (see Table I).

In the preprocessing stage, the infant's face was detected in each frame and 68 facial points were extracted. These points were then used to align the face, crop it, and divide it into four regions. To extract pain-relevant features, optical flow vectors were computed between consecutive frames and used to estimate the optical strain magnitudes, which measure the facial tissue deformations occurred during facial expressions. Then, a peak detector was applied to the strain curves to find the maximum strain magnitudes (i.e., peaks) that correspond to facial expressions. For classification, the extracted strain features were used to train different machine-learning classifiers, namely SVM and K-nearest-neighbors (KNN). KNN is a simple, non-parametric algorithm that stores all instances in advance and classifies a new instance based on a similarity measure, usually a distance function. To evaluate the trained model and estimate the generalization performance, 10-fold cross-validation evaluation protocol was applied. KNN achieved the highest overall accuracy (96\%) for classifying infants' facial expressions as pain or no-pain expressions.

Despite optical flow's popularity and efficiency in motion estimation, the violation of smoothness constraint and self-occlusions can cause the optical flow to fail and provide inaccurate flow computations. Another factor that affects the optical flow results are motion discontinuities and illumination variations. To better handle the smoothness constraints, other flow estimation methods (e.g., SIFT flow \cite{liu2011sift}) were proposed. For robust computation of the flow vectors in the presence of illumination variation and occlusion, different methods are discussed in \cite{kim2005robust,kim2005robust,ince2008occlusion}. 

Other motion-based methods were proposed to detect pain expressions from video sequences of adults; we refer the interested reader to \cite{zhu2014pain} for further reading because this paper focuses mainly in infants.

\paragraph{\textbf{Model-based Methods}}
The basic concept of model-based algorithms is to search for the optimal parameters of an object model that best match the model and the input image. Active Appearance Model (AAM) is a well-known model-based algorithm that uses appearance (i.e., combination of texture and shape) for matching a model image to a new image. It is one of the most commonly used algorithms in various applications such as face recognition \cite{edwards1998face}, facial expression recognition \cite{cheon2009natural}, and medical image analysis \cite{beichel2005robust}. To fit AMM model to a facial image, the error between the representative model and the input image should be minimized (i.e., a non-linear optimization problem).

Fotiadou et al. \cite{Fotiadou2014} discussed using AAM in detecting infants' pain expression during acute painful procedure. The presented method adopts the method proposed in \cite{5643167} to analyze adults' pain expression. The database utilized in this work consists of facial expression data for 10 infants hospitalized in the NICU at a local hospital in Veldhoven, The Netherlands. Infants were recorded in four states, namely heel lancing (i.e., acute procedure), diaper change, hunger, and resting/sleeping. All videos were recorded under unconstrained lighting conditions. 

For each video, AAM tracker was applied to track facial landmark points through the video frames. Then, three features were extracted from the tracked face based on AAM parameters. Specifically, SPTS (similarity-normalized shape), SAPP (similarity-normalized appearance), and CAPP (canonical-normalized appearance) were extracted as discussed in \cite{5643167}. SPTS feature vector contains the coordinates of the landmark points after removing all rigid geometric variations; SAPP vector represents the appearance after removing rigid geometric
variations and scale; and CAPP represents the appearance after removing all the non-rigid shape variation. 

A total of 15 videos for 8 infants were used to build the automated discomfort detection system. The videos of the remaining two infants were excluded from further analysis since these videos include severe occlusion caused by large face rotation or moving hands. The proposed system classified infants' facial expression as discomfort or comfort using the extracted features and an SVM classifier. To evaluate the classifier's performance, leave-one-subject-out cross-validation was performed. The result (0.98 AUC) showed that the proposed system can detect discomfort automatically. 

This work has three main limitations. First, the emotional states for each class was not clearly specified. For example, it was not specified clearly if the discomfort class contains only the heel puncture or if it contains heel puncture as well as diaper change and hunger. We believe that the latter two states are different than pain and, thus, should be treated separately. Second, all the experiments were carried out using a person-specific AAM that is constructed specifically for each infant; this can lead to scalability issues in practice. Third, the proposed method requires further investigation on a larger dataset since it was evaluated on a small dataset (8 subjects). 

We refer the reader to other works \cite{ASHRAF20091788,lucey2008improving,5771462,Hammal:2012:ADP:2388676.2388688} that utilize AAM to assess adults' pain from videos of UNBC-Mac Master Shoulder Pain Expression Archive Database (See Section IV.A for database description). 

\paragraph{\textbf{FACS-based Methods}} FACS is a comprehensive system that uses a set of numeric codes to describe the movements of facial muscles for all observable facial expressions. FACS's numeric codes, which represent the facial muscles' movements, are known as Action Units (AUs). Neonatal Facial Coding System (NFCS) is an extension of FACS designed specifically to observe infants' pain-relevant facial movements (see Table I). 

The vast majority of the methods in the field of automatic facial expressions recognition use FACS to detect facial expressions. However, we are not aware of any FACS-based method that is designed specifically to detect infants' facial expression of pain. In different population, Sikka. et al. \cite{sikka2015automated} presented a FACS-based method to describe children's facial expressions of pain. The proposed method was applied to video sequences of 50 children recorded during ongoing and transient pain conditions. A total of 35 infants were Hispanic, 9 non-Hispanic white, 5 Asian, and 1 Native American; The infants' age ranges from 5 to 18 years old and 35\% of the infants were boys. The data were collected over three visits:  1. within 24 hours of appendectomy surgery; 2. one calendar day after the first visit; and (3) at a follow-up visit. The transient pain was triggered by manually pressing the surgical site for 2-10 seconds.
At each visit, facial expressions of the children were recorded using Canon VIXIA-HF-G10 video camera placed in an upright position. Along with the video recording, self-reported rating by the children and by-proxy rating by both a parent and a nurse were collected to get the ground truth labels.

To extract useful features from the recorded videos, the Computer Expression Recognition Toolbox \cite{5771414} (CERT) was used to detect several AUs. A feature selection method was then applied to select fourteen representative AUs (e.g., AU4 brow lower, AU7 lid tighten, and AU27 mouth open) and different statistics (e.g., the mean, 75th percentile, and 25th percentile) were computed for each of these AUs to form the feature vectors. The extracted features were used to build a logistic regression model evaluated using 10-fold cross validation. The binary classification of facial expression as pain or no pain achieved good-to-excellent accuracy with 0.84-0.94 AUC for both ongoing and transient pain. The primary limitation of this work is the restricted light and motion condition. The presented algorithm requires moderate lighting and motion, which might be difficult to accomplish in clinical settings especially in case of infants in the NICU. As for adults, FACS-based methods were presented to detect facial expression related to pain. We refer the interested reader to \cite{LITTLEWORT20091797,BARTLETT2014738} for further presentation and discussion. 

The main challenge of FACS-based methods is the extensive time required for labeling AUs in each video frame to get the ground truth. It has been reported \cite{LITTLEWORT20091797} that a human expert needs around three hours to code one minute of a video sequence. One-way to reduce the cost of labeling is to automatically detect AUs in each frame and use them as labels. Automatic detection of facial action units in real-world conditions is a challenging area of research that is not directly relevant to this review and, thus will not be discussed further. Those who are interested in the automatic detection of AUs are referred to \cite{bartlett2006fully,valstar2006fully} for more information. 

In summary, we presented above several methods for analyzing infants' facial expression of pain and discussed their limitations. As confirmed by the authors through email, the only database of the above-presented works that is available, per request, for research in ''neonatal'' pain assessment is COPE database. None of the above-mentioned methods' code, except \cite{nanni2010local}, is publicly available. \\

\subsubsection{Infant Cry}
Infant cry is a common sign of discomfort, hunger, or pain. It conveys information that helps caregivers to assess the infant's emotional state and react appropriately. Crying analysis can be divided into two main stages: (1) signal processing stage, which includes preprocessing the signal and extracting representative features; and (2) the classification stage.  We classified the existing methods of signal processing stage into: (1) Time-domain methods; (2) Frequency-domain methods; and (3) Cepstral-domain methods (see Figure 1). For better comparison, we summarized the methods of infant cry analysis in Table III.

\paragraph{\textbf{Time Domain Analysis}} Time domain analysis is the analysis of a signal with respect to time (i.e., the variation of a signal's amplitude over time). Linear Predictive Coefficients (LPC) is one of the most common time-domain methods for analyzing sounds. The main concept behind LPC is the use of a linear combination of the past time-domain samples to predict the current time-domain sample. Other time-domain features that are commonly used for infants' sound analysis are energy, amplitude, and pause duration.  

\begin{table*}
\caption{Summary of Automated Methods for Analyzing Infant Cry} 
\label{table:nonlin}
\begin{tabularx}{\textwidth}{*{6}{l}} 
\toprule
\textbf {Ref. \& Year} & \textbf{Database} &\textbf{Category} & \textbf {Extraction Method} & \textbf{Classification} & \textbf{Results} \\
\midrule
\addlinespace
\cite{6176851}: 2012  & \textbf{Subjects:} 120 infants & Time  & Short-time energy & SVM & Accuracy per class: \\ 
 & \textbf{Race:} N/A   & Domain & (STE) and pause & \textbf{Testing protocol:}  & Pain, 83.33\% \\ 
  & \textbf{Age range:} 12-40 weeks & Analysis & duration &  Splitting samples & Hunger, 27.78\% \\ 
 & \textbf{Stimuli:} N/A  &  &  & into  train and test & Wet-diaper, 61.11\%
  \\
  &  \textbf{Data:} 120 samples; 30 Pain, &  & &  & Avg. accuracy:  \\ 
  & 60 hunger, and 30 wet-diaper &  &  &  & 57.41\% \\ \addlinespace
\hline
\\

\cite{Pal1660444}: 2006 & N/A & Frequency & F0 Fundamental &  K-means: Pain, hunger, & Classification accuracy: \\ 
  &  & Domain& frequency and 3 &  fear, sadness, and anger & 91\% \\ 
 &  & Analysis  & first formants & &  \\ 
 \addlinespace
\hline
\\

\cite{FULLER1988251}: 1988 & \textbf{Subjects:} 41 infants & Frequency & Mean value of  & Statistical analysis & Unique spectral   \\ 
  & \textbf{Race:} Caucasian  & Domain  &  spectral energy & (ANOVA)  & characteristics of \\ 
  & \textbf{Age range:} 2 to 6 months old & Analysis &  &  & pain-induced cry \\ 
 &\textbf{Stimuli:} Immunization for pain,  &  &  &  &  \\ 
  & feeding time for hunger, naptime & &  & &    \\ 
    &for fussy, and fondling for cooing  &  &  &   &    \\
    & \textbf{Data:} 109 samples; 16 hunger, 23 &  &  &  &   \\  
     &cooing, 42 pain, and 28 fussy &  &  &  &     \\ \addlinespace
\hline
\\
\cite{pai2016automatic}: 2016  & \textbf{Subjects:} 27 infants & Frequency & LPC and statistics & kNN: &  Average accuracy: \\ 
 &  \textbf{Race:} Caucasian, Hispanic, & Domain & (e.g., mean and std) & Whimper cry & 76.47\% \\ 
  & African american, and Asian
  & Analysis & & Vigorous cry & \\ 
  & \textbf{Avg. age:} 36 gestation weeks  &  &  & \textbf{Testing Protocol:}  & \\ 
  & \textbf{Stimuli:} Immunization and   &  &   & 10-fold cross validation   &     \\
  & and heel lancing   &  &   &  &     \\
   &\textbf{Data:} 34 samples; NIPS score  &  &   &  &     \\
  \addlinespace
\hline
\\
\cite{Petroni205186}: 1995  & \textbf{Subjects:} 16 & Cepstral & 10 MFCC coeff. & Neural Network: & Classification accuracy: \\ 
& \textbf{Race:} N/A  & Domain & & Pain cry vs  &  Pain (92.0\%) \\ 
 &\textbf{Age range:} 2 to 6 months old  & Analysis &  &  no-pain cry (fear \& & No-pain (75.7\%)   \\ 
 & \textbf{Stimuli:} immunization (pain),  &  &   & anger)  & \\ 
 & jack-in-the-box (fear), and head  &  & & \textbf{Testing Protocol:} &  \\ 
  & restraint (anger).  &  &  & 10-fold cross validation  & \\ 
  & \textbf{Data:} 230 cry samples &  &  &  &  \\
  \addlinespace
 \hline
 \\
\midrule

\cite{Barajas-Montiel2006}: 2006  & \textbf{Subjects, race, age, and }  & Cepstral & 16 MFCC coeff. & FSVM: & Average accuracy: \\ 
 & \textbf{stimuli:} N/A  & Domain & Dimensionality & Pain cry & 97.83\% \\ 
  & \textbf{Data:} 1627 samples; 209 pain, & Analysis & reduction (PCA) & Hunger cry & \\ 
 & 759 hunger, and 659 others & &  &  No-pain-no-hunger cry &  \\ 
  & (Data collected and labelled by doctors) &  & & \textbf{Testing protocol:}  \\ 
  &   &  &  & 10-fold cross-validation  & \\ 
\addlinespace 
\hline
\\
\cite{5466907}: 2010  & \textbf{Subjects and race:} N/A  & Cepstral & 12 MFCC coeff.  & Neural Network: & Classification accuracy: \\ 
 &  \textbf{Age range:} newborns to 1 year & Domain & 16 LPCC coeff. & Pain/no-pain & MFCC: 76.2\% \\
  & \textbf{Stimuli:} Immunization (pain)
  & Analysis & & \textbf{Testing Protocol:}  & LPCC: 68.5\%  \\ 
  & and spontaneous emotions  &  &  & Splitting samples to   & \\ 
  & \textbf{Data:} 180 sample; 150 pain  &  &   &train and test  &     \\
  & and 30 no-pain    &  &   &  &     \\
  \addlinespace
\hline
\\
\cite{6176851}: 2012   & Database in $1^{st}$ row & Cepstral & 13 MFCC,  $\Delta$   & SVM: & Accuracy per class: \\ 
 &   & Domain & MFCC, and $\Delta$ & Pain, hunger, and & Pain(30.56\%) \\ 
  & 
  & Analysis & $\Delta$ MFCC & and wet-diaper & Hunger (66.67\%) \\ 
  &   &  &   & \textbf{Testing Protocol:}  & wetdiaper (86.11\%) \\ 
  & \  &  &   & Splitting samples to   &     \\
  &   &  &   & train and test &     \\
  \addlinespace
\hline
\bottomrule
\end{tabularx}
\end{table*}

Vempada et al. \cite{6176851} presented a time-domain method to detect discomfort-relevant cries. The proposed method was evaluated on a dataset consists of 120 cry corpuses collected during pain (30 corpuses), hunger (60 corpuses), and wet-diaper (30 corpuses). We want to note that the paper does not provide information about the stimulus that triggered the pain state nor the data collection procedure. The infants' age ranges from 12 - 40 weeks old. All corpuses were recorded using a Sony digital recorder with sampling rate of 44.1 kHz. In the feature extraction stage, two features were calculated: 1) Short-time energy (STE), which is the average of the square of the sample values in a suitable window; and 2) Pause duration within the crying segment. Part of these features were used to build SVM and the remaining were used to evaluate its performance. The recognition performance of pain cry, hunger cry, and wet-diaper cry were 83.33\%, 27.78\%, and 61.11\% respectively. The average recognition rate was 57.41\%. 

In a different application, time-domain methods were proposed to analyze infant cry for the purpose of diagnosing a specific disease. The keen reader is referred to \cite{EE} for further discussion.

\paragraph{\textbf{Frequency Domain Analysis}} Frequency domain analysis shows the distribution of the signal within specific ranges of frequencies. The fundamental frequency ($F0$) is a well-known frequency domain property that represents the lowest frequency of a periodic signal. It is worth noting that the fundamental frequency and the pitch, which is a subjective phenomenon that represents the brain's perceptual estimation of the fundamental frequency, are used interchangeably in the literature \cite{Dror2010}. According to \cite{boukydis2012infant}, infant cries can be classified based on the fundamental frequency into: 

 \begin{itemize}
  \item Phonated cries that have a smooth and harmonic structure with a fundamental frequency's range of 400 to 500 Hz. 
  \item Dysphonated cries that have less harmonic structure compared to phonated cries. 
  \item Hyperphonated cries with an abrupt and upward shift in pitch (up to 2000Hz). The hyperphonated cries appear to be associated with a painful stimulus, as discussed in \cite{boukydis2012infant}.
 \end{itemize}
Phonated, dysphonated, and hyperphonated fundamental frequency of infants' sounds can be estimated using different methods presented in \cite{Dror2010,1403155}.

Pal et al. \cite{Pal1660444} used the Harmonic Product Spectrum (HPS) method to extract the fundamental frequency ($F0$) method along with the first three formants (i.e., $F1$, $F2$, and $F3$) from crying signals of infants recorded during several emotional states (i.e., pain, hunger, fear, sadness, and anger). The paper does not provide any information about the database (e.g., number of subjects, age range, and etc.). Moreover, no information was given about the data collection procedure and the stimuli that triggered those emotional states. After extracting the features, k-means algorithm was applied to find the optimal parameters that maximizes the separation between features of different types of cry. Combining $F0$, $F1$ and $F2$ produced the best clustering and achieved an accuracy of 91\% for pain, 72\% for hunger, 71\% for fear, 79\% for sadness, and 58\% for anger. The high accuracy of pain cry can be attributed to the fact that this type of cry has a distinctive and higher fundamental frequency as compared to other types of cries \cite{Pal1660444}. 

Fuller and Horii \cite{FULLER1988251} presented a Frequency domain method to analyze four types of infant cry:  pain, hunger, fussy, and cooing. The utilized database consists of vocal samples collected from 41 healthy infants (2-6 months old). Pain cry samples (i.e., 42 samples) were recorded during a routine intramuscular immunization. Hunger cry samples (i.e., 16 samples) were recorded prior the infant's usual feeding time. Fussy cry samples (i.e., 28 samples) were recorded during the naptime in an infant that was identified as tired. An infant's response to the mother's soft sound and fondling represented the cooing state samples (i.e., 23 samples). In the preprocessing stage, the collected samples were divided into multiple time segments of 512 data points that receive Hamming weighting before computing the fast Fourier transform. Then, the mean value of the spectral energy levels was computed for each vocal sample and used to perform statistical analysis (ANOVA). The result showed that there is a significant difference between the cooing sound and the other cries (pain, hunger, and fussy). It also showed that the spectral characteristics of pain-induced cry is quantitatively different than the other cries (hunger and fussy). Particularly, the spectral energy distribution of pain cry has significantly less difference between the amplitude of the various frequency locations and maximal amplitude than other cries (hunger and fussy) and cooing. 

Pai et al. \cite{pai2016automatic} presented a spectral method to classify infants' cry as a whimper or vigorous. The database of this work was collected from 27 infants, average age is 36 gestational weeks, hospitalized in the NICU at a local hospital in Tampa, Florida. The audio data were recorded during acute painful procedure (i.e., heel lancing and immunization). Two types of pain cry were recorded, whimper cry (14 samples) and vigorous cry (20 samples). The ground truth labels for the recorded samples were given by trained nurses using NIPS pain scale. To obtain the power spectrum for each sample, Welch's method was applied in 20-milliseconds windows. After getting the spectrum, Linear Predictive Coefficients (LPC) along with other statistics (e.g., mean and standard deviation) were extracted from each sample and used to train kNN. The average accuracy of the classifier, evaluated using 10-fold cross validation, was 76.47\%. 

Another Frequency domain method was introduced in \cite{4212215} to detect hunger and other negative emotional states (e.g., sleepy) that are not relevant to pain.

\paragraph{\textbf{Cepstral Domain Analysis }} The Cepstral domain of a signal is generated by taking the Inverse Fourier transform (IFT) of the logarithm of the signal's spectrum. Mel Frequency Cepstral Coefficients (MFCC) is a common Cepstral domain method that is used to extract a useful and representative set of features (i.e., coefficients) from a sound signal and discard noise and non-useful features. 

One of the first studies to analyze infant cry using MFCC was introduced in \cite{Petroni205186}. The proposed method was applied to a database that consists of 230 cry episodes collected from 16 healthy infants (2 to 6 months old). The crying episodes were recorded during three different stimuli: immunization (pain), jack-in-the-box (fear), and head restraint (anger). The cry signals of fear and anger were combined together to represent the no-pain cry. Prior feature extraction, all episodes were filtered to 8000 Hz using low-pass filter, sampled at 16 kHz, and segmented into 256-sample frames (16 ms) with 50\% overlapping. For each segment, 10 MFCCs were extracted and fed into a neural network as input. The testing protocol was 10-fold cross validation. The highest correct classification rates for pain and no pain classes were 92.0\% and 75.7\% respectively.  

Barajas-Montiel et al. \cite{Barajas-Montiel2006} presented MFCC-based method to classify infant cry as pain cry, hunger cry, and no-pain-no-hunger cry (i.e., sleepy and discomfort) using Fuzzy Support Vector Machine (FSVM). FSVM is an extension of SVM that reduces the effect of outliers by assigning a fuzzy value or weight for each training point rather than assigning equal points as in SVM. The database utilized in this work consists of 1627 cry samples collected and labeled by medical doctors. A total of 209 samples were recorded during pain, 759 samples were recorded during hunger, and 659 samples were recorded during other states such as sleepiness and discomfort. In the preprocessing stage, each cry sample was filtered, normalized, and divided into segments of one second. Every one second segment was further divided into frames of 50-milliseconds and 16 MFCC coefficients were extracted from each frame. This procedure generated, for each sample, a high-dimensional vector; PCA was used to reduce the vector dimensionality. FSVM classifier, which was evaluated using 10-fold cross validation, achieved 97.83\% accuracy. 

Yousra and Sharrifah \cite{5466907} introduced a Cepstral domain method to classify infant cry as pain or no-pain (i.e., hunger and anger). A set of 150 pain samples and 30 no-pain samples were recorded for infants with age ranges from newborns up to 12 months old. The pain samples were recorded during routine immunization procedures in a NICU at a local hospital. No-pain samples were recorded at infants' home. Of the 180 recorded samples, 881 samples were obtained by creating one second segments. These samples were then used to extract two sets of features, namely Mel Frequency Cepstral Coefficients (twelve MFCC coefficients) and Linear Predication Cepstral Coefficients (sixteen LPCC coefficients). The extracted features were fed to a neural network trained with the scaled conjugate gradient algorithm; 700 samples were used for training the network and 181 samples were used for testing. The proposed method achieved 68.5\% and 76.2\% accuracies for LPCC and MFCC respectively. This result suggests that MFCC features outperform LPCC features in detecting infant pain cry.

Similarly, Vempada et al. \cite{6176851} investigated the use of MFCC (i.e., 13 MFCCs, 13 delta MFCCs and 13 delta-delta MFCCs) along with other Time domain features for classifying infant cry as pain, hunger, or wet-diaper. A description of the utilized database was provided earlier (see first row of Table III). Each recorded sample was segmented into 20 milliseconds frame. Then, MFCC was applied to each frame to extract useful features for classification. Part of the extracted features was used to build the SVM model and part was used to evaluate its performance. The average accuracies for pain, hunger, and wet-diaper are 30.56\%, 66.67\%, and 86.11\% respectively. Referring to the results of Time domain features (see first row in Table III), it can be seen that the wet-diaper cry has good accuracy using both Time and Cepstral features. However, pain cry is poorly recognized using MFCC features and hunger cry is poorly recognized using Time features. To improve the overall performance, feature fusion and score fusion of Time and Cepstral domains were performed. The feature fusion achieved 77.78\%, 61.11\%, and 83.33\% accuracies for pain, hunger and wet-diaper. The average accuracies using score fusion for pain, hunger, and wet-diaper are 80.56\%, 75\%, and 86.11\% respectively. This result suggests that the fusion of different domains is a good practice for analyzing infant cry.

To summarize, we grouped the existing methods to analyze infant cry into three main categories: Time domain, Frequency domain, and Cepstral domain. Time domain shows the variation of a signal' s amplitude over time. Frequency domain analysis shows the signal's pitch and range of frequencies and it is commonly used to perform filtering operations since it is easier to determine noise in the frequency domain. Cepstral domain shows well-defined harmonic structures in the signal with strong fundamental frequency component and reduced noise. It has been reported \cite{6176851} that analyzing the signal in different domains (i.e., fusion of different domains) provide better performance and might be the best practice for analyzing infant cries. 

Before we proceed to the next section, we would like to note that none of the above-presented works have their database or code publicly available for research use according to the authors who were contacted through email and our online search in public repositories.  \\ 

\subsubsection{Body Movement}
Infants tend to shake their head, extend their arms/legs, and splay their fingers when they experience pain. Therefore, body movement is considered a main indicator in several pediatric scales (see Table I). Following the same structure, this section should provide a discussion of the existing automated methods that analyze body movement for the purpose of assessing infants' pain. However, because there exists no work, except \cite{Zamzmi2017}, that detects and assesses pain based on analysis of body movement, we decided to present the current methods that analyze infants' movements for disease diagnosis. We believe these methods could be used to assess infants' pain since both applications (i.e., assessment and diagnosing) involve measuring spontaneous movements of infants. In fact, a recent work \cite{Zamzmi2017} showed that utilizing the Motion Image, which was introduced in \cite{ADDE2009541} to predict movement disorders in infants, can be used to assess infants' pain effectively. 

We divided the existing methods that analyze infants' body movements for disease diagnosis into Instrument-based and Video-based. We briefly reviewed both categories in the next subsections. A comprehensive survey of the current automated methods for assessing spontaneous general movements in infants can be found in \cite{marcroft2014movement}.

\paragraph{\textbf{Instrument-based Analysis}} Instrument-based methods measure the body movements using specific instruments (e.g., accelerometers and motion sensors) attached to the infant's skin. 

Meinecke et al. \cite{MEINECKE2006125} proposed a kinematic biomechanical model to predict the possibility of developing spasticity, which is a muscle control disorder. Twenty-two infants, fifteen healthy and seven at-risk, took part in this study; the infants' average age is 28.6 gestational weeks. The condition of "at-risk" infants was determined through ultrasound-based detection of cerebral haemorrhage. To segment the infant's body, passive markers were placed in the infant's hands, forearms, upper arms, head, trunk, thighs, lower legs, and feet during spontaneous motor activity. Then, a 3D motion analysis system with a temporal resolution of 50 HZ and a high spatial precision was applied to capture the motion data of these markers. For classifying infants into healthy or at-risk, 53 quantitative parameters were computed from the motion data and used with a Multiple Discriminant Analysis (MDA) method to build a prediction model. The proposed method achieved an overall accuracy of 73\%, a sensitivity of 100\%, and a specificity of 70\%. 

Another instrument-based method was presented in \cite{conover2003using} to diagnose the Cerebral Palsy disorder, which is a motor disorder that affects the infant's ability to move in a coordinated and purposeful way. To measure the body motion, four accelerometers were placed in the infant's arms and legs. Then, Fourier transform was applied to the motion data to analyze the signals in the frequency domain. Next, Multidimensional Cross-correlation technique was performed to study the correlation between different body parts (e.g., correlation between left and right arms). The results showed that acceleration measurement devices can be used for monitoring infants' movement. 

The main advantage of instrument-based methods is the high temporal resolution that allows for detailed in-depth analysis of subtle movements. However, methods under this category require high setup effort and they are obstructive, expensive, and not suitable for clinical applications.

\paragraph{\textbf{Video-based Analysis}}
Video-based methods analyze infants' body movement in raw videos recorded using regular RGB cameras, web-cameras or video-enabled monitors. These methods are more suitable for continuous assessment of infants' movement in clinical environments and homes since they are inexpensive, contactless, and require less setup effort compared to instrument-based methods.  

Different motion estimation methods such as Optical Flow and Motion Image were utilized to estimate infants' body movement from videos. For example, Stahl et al. \cite{stahl2012optical} presented an Optical Flow based algorithm to predict infants at the risk of developing Cerebral Palsy (CP) disorder. The utilized database consists of 136 videos recorded for 82 infants (15 diagnosed with CP and 67 healthy) in the age range of 10-18 weeks. For each video, Optical Flow was applied to generate motion trajectories. Then, these trajectories were transferred to time dependent signals and were analyzed further to extract three types of features: Wavelet Coefficients, Absolute Motion Distance, and Relative Frequency features. The Wavelet Coefficients measure the variety of infants' movements. The other two features measure the activity and the occurring frequencies in the movement patterns. For classification of infants into impaired or unimpaired, linear SVM with 10-fold cross-validation achieved 93.7 $_{-}^{+}$ 2.1, 91. 7 $_{-}^{+}$ 2.2, and 84.7 $_{-}^{+}$ 1.8 average accuracies when it is trained using Relative Frequency features, Absolute Motion Distance, and Wavelet Coefficients respectively.

Instead of applying the Optical Flow in the entire body region, tracking the motion of each body part (i.e., arms, legs, head, and torso) can give a better understanding of the movement pattern for a specific body part and allows studying the correlation between different parts. 

Rahmati et al. \cite{6944446} introduced an Optical Flow based method to classify infants as healthy or at-risk of developing Cerebral Palsy (CP) disorder. Videos were collected from 78 infants (age range of 10-18 weeks); 14 infants were diagnosed with CP and the remaining infants were healthy. The proposed method consists of three main stages: (1) body motion segmentation using optical flow, (2) features extraction, and (3) classification. 

The motion segmentation stage involves three steps. First, it generates a dense trajectory field to track points of the infant's body parts using optical flow. Second, it applies a graph-cut optimization algorithm to separate similar trajectories into different segments (e.g., head segment and left-hand segment). Third, it computes a single representative trajectory for each body part. The generated trajectories are then used to extract three types of features: correlation between trajectories, area out of standard deviation (STD) from moving-average, and periodicity. The correlation between trajectories feature measures the dependencies between the limbs' motions, STD represents the trajectory's deviation from its motion average, and periodicity measures the smoothness/frequency in the movement pattern.  

To classify the infants as healthy or affected, the extracted features were used to train SVM. The average accuracy of evaluating SVM on an unseen dataset (i.e., leave-one-subject-out cross validation) was 87\%; the sensitivity and specificity were 50\% and 95\% respectively. The low sensitivity is attributed to the relatively small numbers of the positive class (i.e., 14 out of 78 infants are at-risk of developing CP). The proposed method can capture small motions and it is robust against noise since it segments the body into separate parts and tracks each part based on a large set of points on that part. 

The Motion Image is a simple and computationally efficient method for motion estimation. It is generated by computing the absolute frame difference between consecutive video frames followed by thresholding to remove noise and get the binary image.  Each pixel in the motion image has a value of zero (i.e., the pixel does not move) or one (i.e., the pixel does move). 

Adde et al. \cite{ADDE2009541} presented a Motion Image based method to classify infants as normal or at-risk of developing CP. The utilized database consists of 136 videos recorded for 82 infants (age range of 10-18 weeks). The presented method started with computing the Motion Image for each video frame. Then, the computed images were used to generate a motiongram by calculating the average of the Motion Image's rows/columns and plotting them over time; averaging over rows shows the horizontal motiongram while averaging over columns shows the vertical motiongram. After generating the motiongram, eight quantitative measures (e.g., quantity of motion) were extracted and used to build Logistic Regression model. The proposed method achieved 81.5\% sensitivity and 70\% specificity. This result shows the feasibility of utilizing the Motion-image as a method for predicting infants' movement disorders.

Zamzmi et al. \cite{Zamzmi2017} adopted the method presented in \cite{ADDE2009541} for pain assessment. The database of this work consists of eighteen infants videotaped in baseline and while experiencing routine acute painful procedures (e.g., heel lancing). Age of the infants was 36 [32, 41] (avg. [min, max]) gestational weeks. All videos were collected and labeled by medical professionals. To classify the emotional state of an infant into pain or no pain, the Motion Images between consecutive video frames were computed. Then, the amount of motion in each frame was computed by summing up the Motion Image's pixels (i.e., total motion per frame). Performing thresholding on the computed total motion feature achieved 87.5\% accuracy. 

To summarize, we divided the existing works that analyze infants' body movements for disease diagnosis into two categories: instrument-based and video-based. Instrument-based methods are more sensitive in capturing the motion and provide high 3D tracking accuracy and resolution. However, these methods are invasive and they are not user friendly, which make them only suitable for research and lab environment. Video-based methods can be easily adapted in clinical environments and homes because they are inexpensive, non-invasive, and require less setup effort.

\subsection{Physiological Measures Based Pain Analysis}
Physiological measures based pain analysis can be defined as the process of extracting pain-relevant features from physiological responses of infants' body. Examples of the physiological responses include changes in vital signs (e.g., increase in heart rate) and cerebral hemodynamic activity (see Figure 1). We discussed in this section the existing methods that utilize physiological measures for pain assessment and summarized them in Table IV.

\subsubsection{Vital Signs Analysis}   
Vital signs readings represent the changes in the body's basic functions such as changes in the heart rate. Caregivers monitor these signs at frequent intervals to check the body condition and understand underlying medical problems. The four main vital signs that are frequently checked by health professionals are Heart Rate (HR), Respiratory Rate (RR), Blood Oxygen Saturation (SpO2), and Blood Pressure (BP). 

The adhesive electrodes and sensors are the most common technology for monitoring infants' vital signs in the NICU. These sensors are placed on the infant's skin to record vital signs signals. Then, the recorded signals are transferred via a translating component to a format that can be displayed on the monitor. To further analyze vital signs data, most monitors provide a wireless data stream to an electronic medical record or allow exporting these signs as time-stamped excel file. 

Different studies utilized vital signs data to study the association between these signs and pain. For example, Lindh et al. \cite{LINDH1999143} described a method to study the association between infants' heart data and acute pain by analyzing the Heart Rate Variability (HRV) in the frequency domain. Vital signs monitor was used to collect heart data from 25 infants (postnatal age of 72--96 hours) in four different events: 1) baseline; 2) sham heel prick (i.e., warming the foot and lancing it with intact lancet); 3) sharp heel prick; 4) and squeezing the heel for blood sampling. The recorded data were inspected for error detection and the artifact were removed by applying interpolation. Then, Statistical and Spectral analyses were carried out on the exported heart data to compute the Heart Rate mean $(HR_{mean})$, the Power in Low Frequency $(P_{LF})$) the Power in High Frequency $(P_{HF})$, and the Total Heart Rate Variability $(P_{tot})$. The computed values were used to perform Multivariate Statistics to illustrate the correlation between these variables and each of the four events. The results showed significant increases in $HR_{mean}$, $P_{tot}$, and $P_{LF}$ between baseline and sharp prick. The results also showed that squeezing the heel for blood sampling during the heel lancing causes a significant increase in $HR_{mean}$ and decrease in $P_{tot}$ and $P_{HF}$ as compared with baseline and sharp prick. 

Faye et al. \cite{faye2010newborn} presented a method to analyze the Heart Rate Variability (HRV) for 28 infants (age $>$ 34 gestational weeks) with chronic pain. EDIN pain scale \cite{faye2010newborn} was used to score the pain and separate infants into two groups: (1) ''Low EDIN'' with EDIN pain score $<$ 5, and (2) ''High EDIN'' with EDIN pain score $\geq$ 5. To study the association between chronic pain and cardiovascular data, Linear Regression Analysis was performed using the mean of Heart Rate $(HR_{mean})$, Respiratory Rate $(RR_{mean})$, Blood Oxygen Saturation $({SpO_{2}}_{mean})$, and High Frequency Variability Index $(HFVI)$. The results showed that HRV changed (i.e., significant decrease) between the two groups; and no significant changes in $RR$ and $SpO_{2}$ were found between the two groups. The results also showed that HFVI ($<$ 0.9 threshold) was able to assess pain with a sensitivity of 90\%, a specificity of 75\%, and 0.81 Area Under the ROC curve. 

Although measuring vital sign using the readily available adhesive electrodes/sensors is the current standard for collecting these data, this standard is expensive, causes stress, and can damage the infants' delicate skin. Therefore, it has been suggested to use contactless and non-invasive methods for monitoring infants' vital signs. Examples of video-based vital signs detection methods are presented in \cite{6226654,AARTS2013943,klaessens2014development,Kumar:15,Poh:10,Balakrishnan_2013_CVPR}. 

In summary, we presented in this section the current efforts for assessing infants' pain using vital signs. Although studies have found a correlation between changes in vital signs and pain, vital sign changes can be associated with other no-pain emotions (e.g., hunger and fear) or underlying illness \cite{bellieni2012pain}. Therefore, it has been suggested \cite{bellieni2012pain,arbour2010vital} to use vital signs in conjunction with behavioral indicators, which are considered more pain-specific, for pain assessment. \\

\begin{table*}
\caption{Summary of Publications for Pain Analysis using Physiological Measures} 
\label{table:nonlin}
\begin{tabularx}{\textwidth}{*{6}{l}} 
\toprule
\textbf {Ref. \& Year} & \textbf{Measures} & \textbf{Database} & \textbf {Extracted Data} & \textbf{Analysis Method} & \textbf{Results} \\
\midrule
\addlinespace
\cite{LINDH1999143}: 1999 & Vital Signs & \textbf{Subjects:} 25 infants  & $HR_{mean}$, the power   & Multivariate  & Increase in $HR_{mean}$,  \\ 
 &   & \textbf{Age range:} 72 - 96 hours & in low frequency ($P_{LF}$), & statistics &  $P_{tot}$, and $P_{LF}$, \\ 
  &  & \textbf{Stimuli:} baseline, sham heel  & and high-frequency ($P_{HF}$)  &  & between baseline and  \\ 
 &  & prick, sharp heel prick, and & and total heart rate variability & & sharp prick
  \\
  &   & heel squeezing & ($P_{tot}$) &  &   \\ 
 \addlinespace
\hline
\\

\cite{faye2010newborn}: 2010 & Vital Signs & \textbf{Subjects:} 28 infants  & Heart Rate Variability& Linear & Sensitivity (90\%)  \\ 
 &   & \textbf{Age:} $>$ 34 gestational weeks & Index (HRVI) & regression & Specificity (75\%) \\ 
  &  & \textbf{Stimuli:} baseline and a major  &  & analysis  & Area under ROC (0.81) \\ 
 &  & surgery (postoperative)  &  & & 
  \\
 \addlinespace
\hline
\\

\cite{BARTOCCI2006109}: 2006 & Cerebral & \textbf{Subjects:} 40 infants, half male & Difference of concentration  & Student t-test & [$HbO_{2}$] increases  \\ 
  & Hemo- & \textbf{Age:} $\geq$ 26 gestational weeks  & of oxygenated [$HbO_{2}$] and & ANOVA  & in both hemispheres; \\ 
  & dynamics & \textbf{Stimuli:} Baseline, tactile, and  & de-oxygenated [$HbH$] and & NewmanKeuls  & more pronounced   \\ 
 & (NIRS) & venipuncture pain stimulus & total ($HbH + HbO_{2}$) hemo- & post hoc test & increase in male \\
  &  & & globin from baseline   &  &    \\ 
 \addlinespace
\hline
\\

\cite{Slater3662}: 2006  & Cerebral & \textbf{Subjects:} 18 infants & Vital signs data and & Statistical t-test & Significant increase \\ 
 &  hemo-  & \textbf{Age:} 25 - 45 postmentsural  & mean of [$HbO_{2}$], [$HbH$], &  & in [$HB_{total}$];   \\ 
  & dynamics & weeks & and $HB_{total} = HbH + $  &  & more pronounced \\ 
  & (NIRS)  & \textbf{Stimuli:} Baseline and heel  & $HbO_{2}$ &   & increase in awake \\ 
  &    & lancing  &   &    & infants    \\
  \addlinespace
\hline
\\
\midrule

\cite{ranger2013multidimensional}: 2013  & Cerebral  & \textbf{Subjects:} 40 infants & $[HbH]_{mean}$,  & Univariate & $\Delta HbH$ differed \\
 & hemo-  & \textbf{Age:} $<$ 12 months & $[HbO_{2}]_{mean}$,  & linear & significantly between \\ 
  & dynamics & \textbf{Stimuli:} Baseline ($T_{0}$), tactile & and $[HR]_{mean}$  & regression  & $T_{0}$ and $T_{2}$\\ 
 & (NIRS) & ($T_{1}$), and painful ($T_{2}$)    &  &   &  \\ 
  &  & stimuli  & &  \\ 
\addlinespace 
\hline
\bottomrule
\end{tabularx}
\end{table*}

\subsubsection{Cerebral Hemodynamics Analysis }
Studies \cite{RANGER2014519,ranger2013multidimensional,BARTOCCI2006109,slater2008well} have shown that there is an association between changes in cerebral oxygenation and pain. The most popular methods to measure the cerebral oxygenation changes are Functional Magnetic Resonance Imaging (fMRI) and Near Infrared Spectroscopy (NIRS). fMRI is a safe method for measuring the brain hemodynamic activity. It produces an activation map that shows which parts of the brain get activated during an emotional event such as pain. NIRS is similar to fMRI but it is less invasive and more suitable for bedside monitoring. It measures, using small probes attached to the head, subtle changes in the concentration of oxygenated hemoglobin [$HbO_{2}$] and de-oxygenated hemoglobin [$HbH$].

Bartocci et al. \cite{BARTOCCI2006109} introduced a NIRS-based method to measure the brain hemodynamic activity for 40 infants, 20 females with age $\geq$ 26 gestational weeks, during three periods: 1) baseline ($P_{0}$); 2) tactile stimulus for cleaning ($P_{1}$); and 3) venipuncture painful stimulus ($P_{2}$). All the data were collected in the NICU at local Hospitals in Sweden and Italy using a double-channel Near Infrared Spectroscopy Device (NIRO 300). This device is widely used in neonatal research to measure functional activation of the cortex.

Each infant was recorded in the baseline period ($P_{0}$) when s/he was in a quiet, awake, and stable condition. The tactile stimulus period ($P_{1}$) was recorded after the disinfecting of an the infant's skin with an alcohol-soaked cotton at room temperature. The painful period ($P_{2}$) was recorded for at least 60 seconds following the insertion of the needle. For all the 40 infants, NIRS data (i.e., $HbH$, $HbO_{2}$, and $HB_{total} = HbH + HbO_{2}$) along with vital signs data (i.e., $HR$ and $SaO_{2}$) were collected during the three time periods. The collected data were sampled and exported to a computer for further analysis. Next, $[HbO_{2}]_{dif}$, $[HbH]_{dif}$, and $[HB_{total}]_{dif}$ were computed by subtracting the values in $P_{0}$ from their values in $P_{1}$ and $P_{2}$ periods. Also, the average of these measurements were computed and used to perform Student's t-test, ANOVA, and Newman–Keuls post hoc statistical tests. The results showed a significant increase in $HR$ and decrease in $SaO_{2}$ between $P_{0}$ and $P_{2}$ periods. For the NIRS measurements, a significant increase was found in the $HbO_{2}$ concentrations in both hemispheres between $P_{0}$ and $P_{2}$ periods; $HbO_{2}$ increase was more pronounced in male than female infants.

Another NIRS-based method was presented in \cite{Slater3662} to measure the brain hemodynamic activity for eighteen infants in the NICU at University College London Hospital, London. The infants' age ranges from 25 to 45 postmenstrual weeks. Vital signs readings along with NIRS data (i.e., $HbH$, $HbO_{2}$, and $HB_{total} = HbH + HbO_{2}$) were recorded, using NIRO 300 device, during baseline and heel lancing periods. The data collection of baseline was performed 20 seconds pre-stimulus. After the insertion of the lancet, the infant's foot was not squeezed for a period of 30 seconds to ensure that the evoked response occurred because of the initial stimulus not the squeezing. The collected data were sampled and the maximum changes from the baseline were calculated for each measure. The result of the statistical analysis (t-test) indicated that the painful stimulus produced a clear cortical response that is measured as an increase in $HB_{total}$ in the contralateral somatosensory cortex. This cortical response was more pronounced in awake infants than in sleeping infants. Moreover, it has been found that the response in the contralateral somatosensory cortex for awake infants increases with age.

Extensions of this work are presented in \cite{slater2008well} to study the relation between NIRS data and behavioral indicators of pain and in \cite{SLATER2010583} to investigate the impact of age and frequency of painful procedures on the brain neuronal responses.

For chronic pain, Ranger et al. \cite{ranger2013multidimensional} presented a NIRS-based method to assess infants' chronic pain based on analysis of hemodynamic activity in brain regions. NIRS data (i.e., $HbO_{2}$ and $HbH$) for forty infants ($<$12 months) were recorded, using NIRO 300 device, during the following periods: 1) chest-drain removal procedure following cardiac surgery ($T_{2}$); 2) removal of the dress ($T_{1}$);  3) and baseline ($T_{0}$). To verify associations between NIRS data and pain stimulus, Univariate Linear Regression was performed on the extracted measures. The results showed a significant increase in $ \Delta HbH$ during pain (i.e., the difference of $  \Delta HbH$ measurement between the baseline ($T_{0}$) and pain ($T_{2}$) was significant). 

In a different population (i.e., adults), fMRI analysis was performed during baseline and different events of thermal stimulation. We refer the reader to \cite{MARQUAND20102178,brown2011towards} for further readings.

As a final remark, we would like to draw the reader's attention to the difference of cortical response between chronic and acute pain. Acute pain produced changes measured as an increase in $HbO_{2}$ \cite{BARTOCCI2006109} or $HB_{total}$ \cite{Slater3662} while the chronic pain caused an increase of $HbH$ \cite{ranger2013multidimensional}. To summarize, we discussed above the current efforts for assessing infants' pain based on analysis of cerebral hemodynamic activity. Although different studies found a strong association between pain and the cortical responses, this area of research is recent and requires more investigation and validation. 

Before we proceed to the next section, we note that none of the above-described databases are publicly available according to the authors, contacted via email, and our online search in public repositories.

\subsection{Multimodal Pain Detection} 
The methods discussed so far utilize a single indicator or modality for detecting infants' pain. Because pain is expressed through multiple modalities, existing pediatric pain scales are multimodal incorporating both behavioral and physiological indicators for assessment. Multimodality allows for a reliable assessment of pain in case of missing data due to occlusion, noise, gestational age (e.g., weak facial muscles in premature infants), physical exertion (i.e., exhaustion), or sedation. In this section, we present the existing automated multimodal methods for assessing pain. We divided these methods based on the fusion level into: Decision-level and Feature-level (see Figure 1). \\

\subsubsection{Decision-level Fusion} 
The decision-level fusion represents a variety of methods designed to merge the decisions or outcomes of multiple classifiers into one single ensemble decision. In other words, decision-level methods take into account the outcome of multiple classifiers, one classifier per pain indicator or modality, to determine the final decision or outcome. Several approaches have been proposed \cite{5871582} to combine different modalities' outcomes for decision-making. 

Majority voting is one of the most common approaches to combine the outcomes of different modalities. In the majority-voting scheme, each indicator contributes one vote (i.e., class label) and the majority label in the combination is chosen as the final decision or outcome. Zamzmi et al. \cite{7900284} utilized a majority voting method to combine different pain indicators for the purpose of developing a multimodal pain assessment system.  Specifically, facial expression, body movement, and vital signs data were collected from 18 infants (average age is 36 gestational weeks) during acute painful procedures and used to extract pain-relevant features. The extracted features for each indicator (i.e., facial expression, body movement, and vital signs) were used individually to build a classifier. For example, the extracted features from facial expression were used with SVM classifier to classify the infant's state as no-pain, moderate pain, or severe pain. To create a multimodal assessment, the outcome (i.e., no-pain, moderate, or severe labels) of the facial expression classifier was combined with outcome labels of body movement and vital signs classifiers. Then, a majority voting method was applied to the combination of different pain indicators. Utilizing the majority-voting method to combine different pain indicators achieved around 95\% accuracy for a combination of vital signs readings, facial expression, and body movement. This work was extended in \cite{Zamzmi2017} to include infant cry to facial expression, body movement, and vital signs readings.

Pal et al. \cite{Pal1660444} described a multimodal emotion detection method that predicts the infant's emotional state based on analysis of facial expression and crying. Geometric features were extracted from the infant's facial expression and the fundamental frequency along with the first three formants were extracted from the crying signals. The extracted features for each modality were then used to build a classifier and a decision-level fusion method was applied to combine the decision labels for both classifiers. Specifically, facial expression and crying modalities were combined by finding the conditional probability matrixes and using the index for the maximum value of a belief vector, which is derived from the probability matrixes, as the final fused decision. The overall accuracy for predicting infants' emotions using a decision-level fusion method was 75.2\%.

One of the main advantages of decision-level fusion is the easy implementation because it depends on combining different classifiers' labels. However, this level of fusion can result in loss of information (i.e., loss of correlation information between indicators) because it depends on the assumption that the combined indicators/modalities are independent. We discuss below another level of fusion that mitigates this issue. \\

\subsubsection{Feature-level Fusion} 
Feature-level fusion is the process of combining multiple modalities in the early stage by concatenating features of all modalities into a single high-dimensional feature vector. The concatenated feature vector is then used to train a single classifier for classification. Theoretically, feature-level fusion can have higher performance than decision-level fusion since it contains much richer information. However, this level of fusion can raise several issues in practice and inappropriate handling of these issues might decrease performance. For example, concatenating the features of different pain modalities into single high-dimensional feature vector may lead to the curse of dimensionality; dimensionality reduction methods such as PCA could be applied in this case to reduce the dimensions. Another issue of feature-level fusion is the missing data due to the failure of recording a specific modality or the unavailability of data at a specific time. Several works proposed methods for handling data that are partially or completely missing. We refer the interested reader to \cite{5871582} for more discussion about fusion methods of multimodal systems with missing data.

To the best of our knowledge, there is currently no work that combines different pain indicators at the feature level for the purpose of assessing infants' pain. In a different population, Werner et al. \cite{6977497} presented the first work that combines video and biomedical signals for pain assessment in adults. The utilized database consists of video and biomedical signals (i.e., Electrocardiogram [ECG], Galvanic Skin Response [GSR], and Electromyography [EMG]) collected for 90 subjects undergoing heat stimulus\footnote{Description of this database (BioVid) is provided in Section IV.A}. To extract pain-relevant features from videos, facial landmark points were detected in each frame and head pose was estimated. These points were used to compute several geometric features (distances) and gradient-based features. Then, the frames were grouped into time windows of 5.5 seconds to form a temporal descriptor or vector for classification. For the biomedical signals, all the signals were divided into windows of 5.5 seconds and filtered to remove noise using a Butterworth bandpass filter. To extract features for classification, different statistics (e.g., mean and standard deviation) were computed from the filtered biomedical signals. In the final step, the features extracted from both video and biomedical signals were fused together to form a single high-dimensional vector, which was used to train random forest model. The proposed method achieved up to 80.6\% mean accuracy for a fusion of video and biomedical signals. 

To summarize, we discuss above two levels, namely decision fusion and feature fusion, for combining different pain indicators. Decision-level fusion assumes that the modalities are independent and ignores the correlation between them. Feature-level fusion can mitigate this issue by combining all the modalities together in a rich and high-dimensional feature vector. However, the high-dimensionality of the feature vector along with the scaling and missing data can raise several issues in practice. These issues can be handled using methods such as standardization (i.e., z-scores) for scaling, PCA for reduction, and interpolation for the missing data. 

\section{Pain Recognition} 
We divided pain recognition into two main classification tasks: pain detection and pain intensity estimation. We present next a description and a discussion of limitations for each task.  

\subsection{Pain Detection}
Pain detection aims to identify the presence or absence of pain emotion. It is a typical classification problem in which discrete classes are considered the output of a classifier. For example, a classifier that is trained with pain-relevant features can be used to classify the emotional state of an infant as pain or no-pain. 

SVM classifier is commonly used for pain detection (e.g., facial expression \cite{LITTLEWORT20091797, ASHRAF20091788,wei2011pain,Sikka2014, werner2013towards, niese2009towards, LITTLEWORT20091797, 7349412, BRAHNAM2006211, nanni2010local, zamzami2015pain, Fotiadou2014, BRAHNAM20071242, Brahnam2007}, cry \cite{6176851, pai2016automatic, Barajas-Montiel2006}, and body movement \cite{stahl2012optical,6944446}). Other classifiers that are used for pain detection are Neural Network \cite{4042206,Petroni205186, Brahnam2007}, k-nearest neighbors \cite{pai2016automatic,zamzami2015pain}, and k-means \cite{Pal1660444}. Such classifiers achieved varying levels of performance in detecting the pain label. 

Pain detection provides the pain label, but does not provide the intensity or level of the detected pain. For pain assessment application, detecting the pain without its intensity may not be enough due to three main reasons. First, providing the pain label without its level does not reflect the severity of pain. Second, it does not reflect the individual differences in response to painful stimuli. Third, producing the label without its intensity does not provide information about the pain dynamic and how it changes over time; an infant might experience different pain intensities at different time intervals. For these reasons, we believe estimating the intensity of pain is important and can lead to better understanding and intervention.

\subsection{Pain Intensity Estimation}
Estimating the intensity of the detected provides better pain assessment and might lead to better pain management. 

Several pain recognition methods were proposed for pain intensity estimation. For example, Gholami et al. \cite{5415598} presented a method (see Table II, 3rd row) to estimate pain intensity using RVM. Unlike SVM, RVM classifier outputs the probabilities of the class memberships or labels. The uncertainty for each class membership was used to estimate infants' pain intensity. For validation, the automated intensity estimation was compared with the intensity estimation provided by experts and non-expert observers. The agreement between RVM and human observers, measured using kappa coefficient, was 0.48 for experts and 0.52 for non-experts. 

Hammal et al. \cite{Hammal:2012:ADP:2388676.2388688} described a method to estimate pain intensities for 25 subjects with an orthopedic injury. Four SVM classifiers were built separately to automatically assess four levels of pain. To measure the reliability of judgments between the automatic estimation and the manual estimation, Intra-class Correlation Coefficient (ICC) that ranges from -1 to 1 was used. The results showed moderate (0.55 ICC) to high (0.85 ICC) consistency between the manual and automated pain intensity assessment.

Similarly, Gruss et al. introduced a method \cite{gruss2015pain} to estimate four levels of pain using SVM. Facial expression and biopotentials signals were recorded under four levels of pain (T1 to T4) as described in Section IV.A (BioVid Heat Pain Database). Then, the recorded signals were analyzed to extract complex mathematical features. These features were used to build SVM classifiers trained with 75\% of the data and tested on 25\% of data. The proposed method achieved 76.00\% (sensitivity) and 82.59\% (specificity) for baseline vs T1, 80.00\% (sensitivity) and 82.59\% (specificity) for baseline vs T2, 84.71\% (sensitivity) and 85.18\%  (specificity) for baseline vs T3, and 92.24\%, (sensitivity) and 89.65\% (specificity) for baseline vs T4. 

\section{Pain Databases} 
The quality, complexity, and capacity are three important factors that should be considered when collecting databases for pain assessment. Low-quality databases with a vague notion of suffering and inadequate annotations can lead to inaccurate results. Also, the complexity of the database, in term of its modalities/dimensions, is critical to develop reliable multimodal pain assessment system that can still assess pain in case of missing data. Finally, databases with relatively small number of subjects are not sufficient to evaluate the system performance and draw conclusions. Therefore, collecting high-quality, multimodal, and large databases is necessary for developing robust pain assessment systems. 

Most of the existing pain databases are not publicly available, due to legal/ethical reasons, for research use. This section provides brief descriptions of the publicly available pain databases for adults and infants. 

\subsection{Adult} 
\textit{UNBC-McMaster Shoulder Pain Expression Archive} \cite{5771462} is one of the first databases that addressed the need for adequately annotated and publicly available database of pain expression. The database consists of videos collected from 129 subjects (63 males and 66 females) during a series of movements to test their affected and unaffected shoulder. All videos were manually coded using FACS (48398 FACS coded frames). In addition to the videos, the database has self-report and observer ratings for each sequence. 

Instead of recording a single moadilty/indicator, Walter et al. introduced \cite{6617456} an advanced and multimodal database, known as the \textit{BioVid Heat Pain Database}. This database contains video and biopotentials signals (i.e., Skin Conductance Level [SCL], Electrocardiogram [ECG], Electromyogram [EMG], and Electroencephalography [EEG]) for 90 subjects with age distributions of 18 to 35 (group 1), 36 to 50 (group 2), and 51 to 65 (group 3). Each group has a total of 30 subjects (15 male and 15 female). All subjects underwent experimentally induced heat stimulus with four intensities or pain levels (T1 to T4). To adjust the level of the stimulation, a subject-specific pain threshold and a pain tolerance were determined. Every pain level was stimulated 20 times (i.e., a total of 80 stimulation). In each stimulus, the maximum temperature the subject can take was held for four seconds and there was a pause duration of 8--12 seconds between the stimuli. This procedure was repeated twice, once when the subject's face was recorded and once when the biopotentials sensors were attached. The subject's face and head pose were recorded using three cameras (AVT Pike F145C cameras) and a Kinect. The biopotentials data were recorded using a Nexus-32 amplifier. More discussion about the experiment setup, sensors' channels, and the synchronization procedure of this database can be found in \cite{6617456}.

\subsection{Infant} 
\textit{COPE/iCOPE}, collected by Brahnam et al. \cite{brahnam2005svm}, is the first pain expression database that is designed specifically to assess infants' pain automatically. The database consists of 204 static images captured, using Nikon D100 digital camera, for 26 healthy infants (50\% female). The infants' age ranges from 18 hours to 3 days old. Before the photography session, all infants were fed and they were swaddled to get an unobstructed image of the face. The images for each infant were taken during four stimuli: 1) the puncture of a heel lance; 2) friction on the external lateral surface of the heel; 3) transport from one crib to another; and 4) an air stimulus to provoke an eye squeeze. The main limitation of this database is the 2D static images that do not show the expression's dynamic and how it evolves over time. Currently, Dr. Brahnam and her collaborators are working on collecting a new and challenging video database (COPE 2). This database is not yet available for research use. Another limitation of this database is the single modality (i.e., facial expression). As discussed in \cite{FRANCK1997343,allegaert2005variability}, incorporating different pain indicators is important to ensure proper and reliable assessment of pain.

Another publicly available neonatal pain database is described in \cite{Harrison2014}. The database consists of \textit{YouTube videos} recorded, by parents or a guardian, for infants receiving immunization injections; the infants' age ranges from less than a month to 12 months old. The recorded videos show the infant's face, body, and have sounds. Along with the raw videos, other data such as the infant's gender, number of injections, and the gender of the caregiver were collected. All videos were scored by experts using FLACC (Face, Legs, Activity, Cry, Consolability) \cite{jaskowski1998flacc} pain scale. The main limitation of this database is the low-quality of the recorded videos which leads to exclude many videos from annotations. 

As far as we are aware, COPE and YouTube databases are the only available neonatal databases for research in pain detection. Therefore, collecting high-quality, multimodal, and relatively large pain databases is needed to advance the automated assessment of neonatal pain. 

\section{Limitation and Future Directions} 
There are several limitations that should be addressed to advance the automated assessment of neonatal pain. These limitations can be summarized as follows: 

\begin{itemize}
\item As discussed above, there are very few accessible databases for research in neonatal pain. At the time of writing this paper, we are only aware of two databases, COPE and YouTube videos, that are available per request for research in neonatal pain assessment. To advance the automated assessment of neonatal pain, researchers need to have access to advanced and multimodal databases that are collected and annotated by experts in the field. 

\item Existing methods for automatic pain assessment focus mainly on adults. We think this focus is attributed, in addition to the database-accessibility issue, to the common belief that the algorithms designed for adults should have similar performance when applied to infants. Contrary to this belief, we think the methods designed for assessing adults' pain will not have similar performance and might completely fail for two reasons. 
First, the facial morphology and dynamics vary between infants and adults as reported in \cite{GRUNAU1987395}. Furthermore, infants' facial expressions include additional movements and units that are not present in the Facial Action Coding System (FACS). As such, Neonatal FACS was introduced and designed specifically for infants. Second, we think the preprocessing stage (e.g., face tracking) is more challenging in infants because they are uncooperative subjects recorded in an unconstrained environment (i.e., NICU).

\item Most of the existing approaches assess pain based on analysis of a single modality (e.g., facial expression). Studies \cite{FRANCK1997343,allegaert2005variability} have shown that pain causes behavioral and physiological changes and suggested considering multiple modalities for better pain assessment. In addition, it has been reported \cite{gibbins2008comparison} that some infants have limited ability to behaviorally express pain due to developmental stage, movement disorder, or physical exertion (e.g., exhaustion or sedation). Therefore, it is important to develop multimodal approaches that can better handle the missing data.  

\item Existing methods for assessing pain do not take the contextual and clinical data (e.g., medication, age, race, and gender) into account when analyzing pain. Studies \cite{gibbins2008comparison,valeri2012pain} found an association between infants' clinical data and their reaction to pain experience. For example, it has been shown \cite{gibbins2008comparison} that infants of different age groups have different pain response. Hence, incorporating clinical and contextual information with other pain indicators is necessary to refine the assessment process and obtain a context-sensitive pain assessment system. 

\item Existing methods for assessing pain focus on the acute procedural pain and neglect postoperative pain, although the continuous assessment and immediate intervention are more needed for the latter.
\end{itemize}

\section{Conclusion}
The current standard for assessing infants' pain is inconsistent and intermittent and needs machine-based techniques to provide consistent and continuous assessment. The automated assessment of pain has three main stages, preprocessing, pain analysis or feature extraction, and pain recognition. This paper presents a comprehensive review of the automated methods for pain analysis and recognition. It also gives descriptions of the databases that are available to researchers, discusses the current limitations of automated pain assessment systems, and suggests directions for future research.


\section*{Acknowledgment} 
Many thanks to the anonymous reviewers whose insightful feedback and constructive suggestions helped in shaping this article into its present form. 

\ifCLASSOPTIONcaptionsoff
  \newpage
\fi



\begin{IEEEbiography}[{\includegraphics[width=1in,height=1.25in,clip,keepaspectratio]{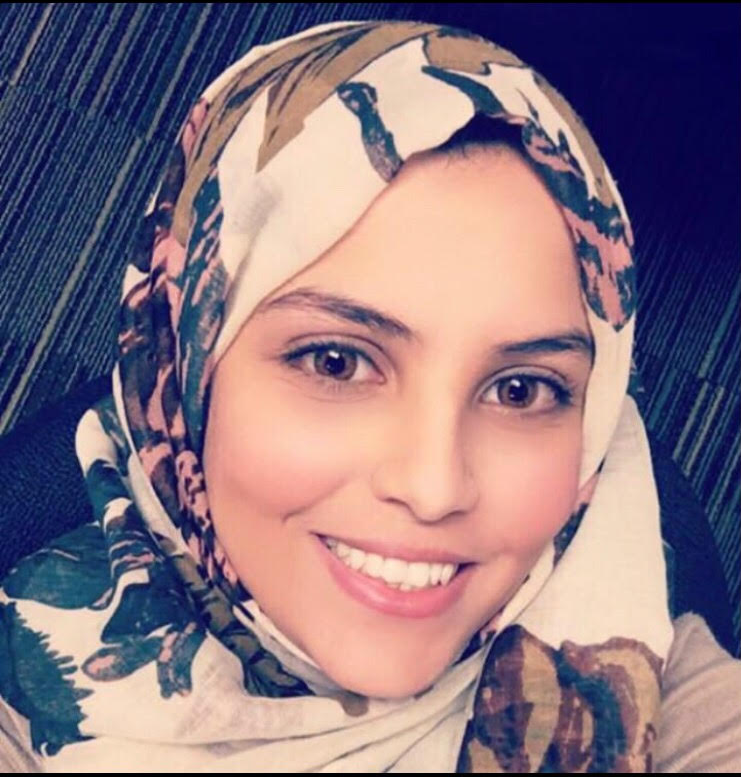}}]{Ghada Zamzmi}
received the M.S. degree in Computer Science from the Department of Computer Science and Engineering at the University of South Florida in 2014. She is currently a doctoral student at the same university.  Her research emphasis is on emotion recognition, in particular, pain recognition for infants and individuals with communicative/neurological impairments.

Her research interests include computer vision, machine learning, emotion recognition, human computer interaction, and medical image analysis.

\end{IEEEbiography}

\vspace{-13 mm}

\begin{IEEEbiography}[{\includegraphics[width=1 in,height=1.25in,clip,keepaspectratio]{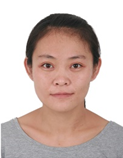}}]{Ruicong Zhi}
received the PhD degree in signal and information processing from Beijing Jiaotong University in 2010. From 2016 to 2017, she visited the University of South Florida as a visiting scholar. She is currently an associate professor in School of Computer and Communication Engineering, University of Science and Technology Beijing. She has published more than 50 papers and has six patents. She has been the recipient of more than ten awards, including the National Excellent Doctoral Dissertation Award nomination. Her research interests include facial and behavior analysis, artificial intelligence, and pattern recognition.
\end{IEEEbiography}

\vspace{-13 mm}

\begin{IEEEbiography}[{\includegraphics[width=1in,height=1.25in,clip,keepaspectratio]{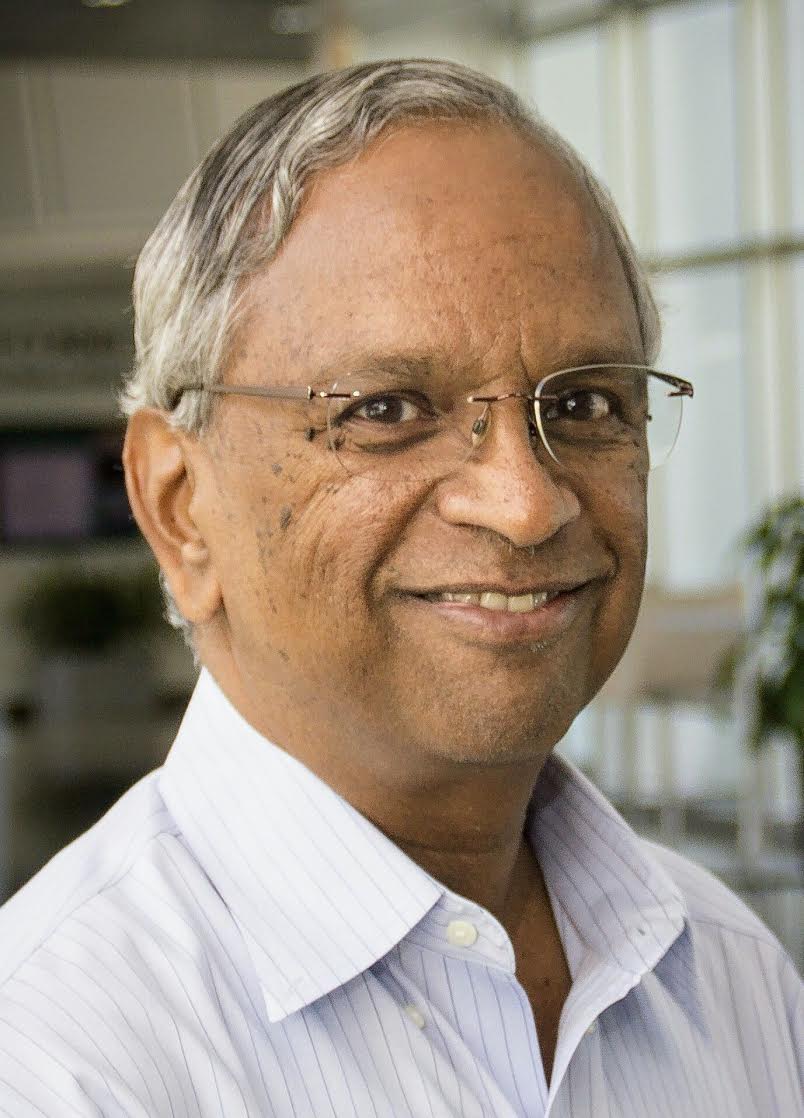}}]{Rangachar Kasturi}
is the Douglas W. Hood Professor of Computer Science and Engineering at the University of South Florida. Earlier, he was a Professor at the Pennsylvania State University.  He received his Ph.D. from Texas Tech University. His research interests are in computer vision, pattern recognition, and document image analysis.  He has served as the Editor in Chief of the IEEE Transactions on Pattern Analysis and Machine Intelligence, a Fulbright Scholar/Specialist, the President of the International Association for Pattern Recognition (IAPR), and the President of the IEEE Computer Society. He is a Fellow of the IEEE and IAPR. 
\end{IEEEbiography}

\vspace{-5 mm}

\begin{IEEEbiography}[{\includegraphics[width=1in,height=1.25in,clip,keepaspectratio]{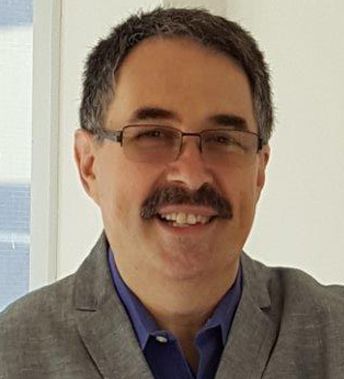}}]{Dmitry Goldgof}
is an educator and scientist working in the area of Medical Image Analysis, Image and Video Processing, Computer Vision and Bioinformatics. Dr. Goldgof is currently Professor in the Department of Computer Science and Engineering at the University of South Florida in Tampa. Dr. Goldgof has graduated 28 Ph.D. and 44 MS students, published over 95 journal and 220 conference papers, 20 books chapters and edited 5 books (h-index 50). Professor Goldgof is a Fellow of IEEE, a Fellow of IAPR, a Fellow of AAAS and a Fellow of AIMBE.
\end{IEEEbiography}

\vspace{-10mm}

\begin{IEEEbiography}[{\includegraphics[width=1 in,height=1.25in,clip,keepaspectratio]{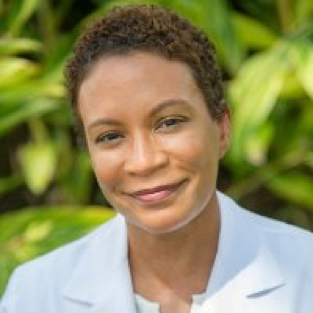}}]{Terri Ashmeade} 
is a Neonatologist and clinical researcher working in the area of clinical care improvement in the Neonatal Intensive Care Unit. Dr. Ashmeade received her medical degree from the School of Medicine at University of Connecticut and has been in practice for more than 20 years. She is currently a Professor in the Department of Pediatrics at the University of South Florida, Morsani College of Medicine in Tampa, Florida. 
\end{IEEEbiography}

\vspace{-10mm}

\begin{IEEEbiography}[{\includegraphics[width=1in,height=1.25in,clip,keepaspectratio]{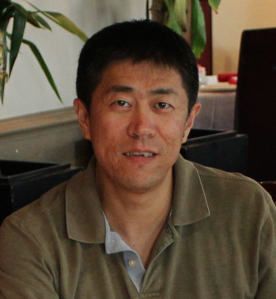}}]{Yu Sun}
is an Associate Professor in the Department of Computer Science and Engineering at the University of South Florida. He is an Associate Editor of the IEEE Transactions on Robotics. He received his Ph.D. degree in Computer Science from the University of Utah in 2007, and his B.S. and M.S. degrees in Electrical Engineering from Dalian University of Technology in 1997 and 2000 respectively. He was a Postdoctoral Associate at Mitsubishi Electric Research Laboratories (MERL), Cambridge, MA from Dec. 2007 to May 2008. His research interests include intelligent systems, robotics, virtual reality, and medical applications. 
\end{IEEEbiography}








\end{document}